\documentclass[10pt]{article} % For LaTeX2e

\usepackage[accepted]{rlj}           % Should be uncommented for submission
\usepackage{amssymb}            % Defines common symbols like \mathbb R
\usepackage{mathtools}          % Extends amsmath, providing common math tools
\usepackage{mathrsfs}           % Enables \mathscr, which can work in cases that \mathcal does not
\usepackage{graphicx}           % For including images
\usepackage{subcaption}         % Allows for the use of subfigures and subcaptions
\usepackage[space]{grffile}     % For spaces in image names
\usepackage{url}                % For displaying URLs
\usepackage{lipsum}             % For placeholder text
\usepackage{booktabs}
\usepackage{wrapfig}
\usepackage{multirow}
\usepackage{ulem} % for underline, replace \ul with \uline

\newcommand{\poke}{\text{Pok\'emon} }
\newcommand{\pokelower}{\text{pok\'emon} }
\definecolor{bo}{HTML}{BF5700}

\title{Human-Level Competitive \poke via Scalable\\Offline Reinforcement Learning with Transformers}
\setrunningtitle{Metamon: Human-Level Competitive \poke}

\author{Jake Grigsby, Yuqi Xie$^\dagger$, Justin Sasek$^\dagger$, Steven Zheng$^\dagger$, Yuke Zhu}

\emails{\{grigsby,yukez\}@cs.utexas.edu}

\affiliations{
\textbf{The University of Texas at Austin}\\
\par % If including additional comments like below, use \par to add some whitespace. 
$^\dagger$ Equal contribution. Order determined by a surprise \poke tournament.
}

\contribution{
    We build and release an offline RL dataset comprising $3.5$M trajectories reconstructed from years of human gameplay in the complex decision-making task of Competitive \poke\hspace{-1mm}.
    }
    {
    PokéChamp \cite{karten2025pokechamp} concurrently released a dataset of \poke battles. The datasets differ in that:
    \begin{itemize}
        \item Ours covers all available data ($2014$-Present) for a smaller list of popular game modes. This provides more demonstrations per mode and explores the challenges of learning from strategies that evolve over time.
        \item Ours is distributed in a flexible RL format that allows for customization of observations, actions, and rewards outside of LLM prompts. 
        \item Ours reconstructs the agent’s partially observed perspective from spectator data with more accuracy thanks to a custom state-tracking and prediction pipeline designed for this purpose. Further discussion is provided in Appendix \ref{app:replay_reconstruction} and in our open-source release. 
    \end{itemize}
  
    }
\contribution{
    We demonstrate our dataset's ability to produce sequence policies that play Competitive \poke at a human level. 
    }
    {
    Prior work has used online self-play and heuristic search to build successful \poke agents in other rulesets.
    }

\keywords{\poke, Offline RL, Imitation Learning} % Your keywords

\summary{Competitive Pokémon Singles (CPS) is a popular strategy game where players learn to exploit their opponent based on imperfect information in battles that can last more than one hundred stochastic turns. AI research in CPS has been led by heuristic tree search and online self-play, but the game may also create a platform to study adaptive policies trained offline on large datasets. We develop a pipeline to reconstruct the first-person perspective of an agent from logs saved from the third-person perspective of a spectator, thereby unlocking a dataset of real human battles spanning more than a decade that grows larger every day. This dataset enables a black-box approach where we train large sequence models to adapt to their opponent based solely on their input trajectory while selecting moves without explicit search of any kind. We study a progression from imitation learning to offline RL and offline fine-tuning on self-play data in the hardcore competitive setting of Pokémon’s four oldest (and most partially observed) game generations. The resulting agents outperform a recent LLM Agent approach and a strong heuristic search engine. While playing anonymously in online battles against humans, our best agents climb to rankings inside the top $10\%$ of active players. All agent checkpoints, training details, datasets, and baselines are available at \href{https://metamon.tech}{\texttt{metamon.tech}}.
}

\begin{document}

%\makeCover  % Create the cover page
\maketitle  % Make the title section

\begin{abstract}
Competitive Pokémon Singles (CPS) is a popular strategy game where players learn to exploit their opponent based on imperfect information in battles that can last more than one hundred stochastic turns. AI research in CPS has been led by heuristic tree search and online self-play, but the game may also create a platform to study adaptive policies trained offline on large datasets. We develop a pipeline to reconstruct the first-person perspective of an agent from logs saved from the third-person perspective of a spectator, thereby unlocking a dataset of real human battles spanning more than a decade that grows larger every day. This dataset enables a black-box approach where we train large sequence models to adapt to their opponent based solely on their input trajectory while selecting moves without explicit search of any kind. We study a progression from imitation learning to offline RL and offline fine-tuning on self-play data in the hardcore competitive setting of Pokémon’s four oldest (and most partially observed) game generations. The resulting agents outperform a recent LLM Agent approach and a strong heuristic search engine. While playing anonymously in online battles against humans, our best agents climb to rankings inside the top $10\%$ of active players. All agent checkpoints, training details, datasets, and baselines are available at \href{https://metamon.tech}{\texttt{metamon.tech}}.
\end{abstract}

\begin{figure}[h!]
    \centering
    \includegraphics[width=.99\linewidth]{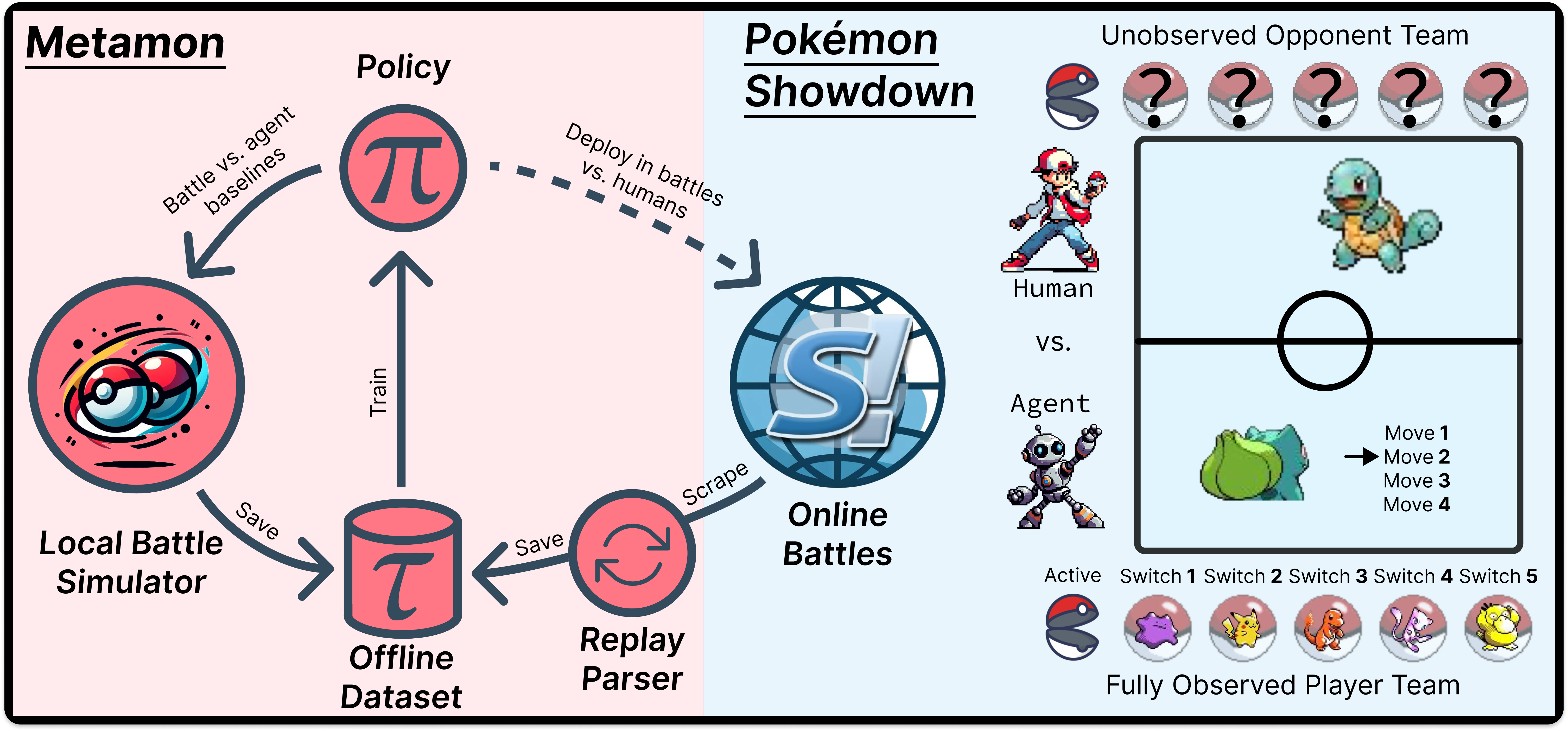}
    \caption{\textbf{Batch Training and Evaluation in CPS.} We develop a platform called \texttt{Metamon} that enables an offline RL workflow on a dataset of human gameplay from \poke Showdown.}
    \label{fig:fig1}
\end{figure}
\section{Introduction}

Competitive \poke (Singles) (\textbf{CPS}) is a two-player strategy game that combines the long planning horizons of chess with the imperfect information, opponent modeling, and stochasticity of poker --- and then adds so many named entities and niche gameplay mechanics that it takes an \href{https://bulbapedia.bulbagarden.net/wiki/Main_Page}{encyclopedia} to document them all. In CPS, players construct teams from billions of possibilities and battle against an opponent. On each turn of the battle, players can choose to use a move from the \poke already on the field or switch to another member of their team (Figure \ref{fig:fig1} Right). Moves can deal damage to the opponent, eventually causing it to faint, until the last player with active \poke wins. CPS AI is an exciting Reinforcement Learning (RL) problem because it requires reasoning under uncertainty in an incredibly large state space. The best \poke AI relies on heuristic search in custom simulators \citep{foul-play} or test-time Monte Carlo tree search with self-play \citep{WinningatPokémonRandomBattles}. Notably, Competitive \poke\hspace{-1mm} is played on a website that saves turn-by-turn records of battles dating back over a decade. We develop a pipeline to convert these logs to the partially observed point-of-view of an agent playing against humans in official ranked battles, thereby unlocking a naturally occurring source of offline RL data \citep{lange2012batch} that grows larger every day. Our ``reconstruction'' process is specific to CPS and will create further CPS-specific problems that RL will need to overcome. At a high level, though, it is an example of a challenge that may arise when using existing data to kickstart a data flywheel. There are applications of RL (healthcare, finance) where lots of data \textit{surrounding} the problem exists (patient records, time series) but is not formatted as trajectory data from the point-of-view of an agent, and any conversion to this format would open up a ``sim-to-real'' gap between the reconstructed (PO)MDP and the real world.

Our dataset enables a general perspective on the CPS AI problem that has previously been impractical: that sequence models might be able to learn to play without explicit search or heuristics by using model-free RL and long-term memory to infer their opponent's team and tendencies. Our experiments take this perspective to its extreme and create a case study in the process of training and evaluating large policies (Fig. \ref{fig:fig1} Left). We develop a suite of heuristic and imitation learning (IL) opponents for offline evaluation with procedurally generated \poke teams. With these opponents as a benchmark, we evaluate Transformers \citep{vaswani2017attention} of up to $200$M parameters trained by IL and offline RL. When deployed in ranked battles against human players in the highly competitive realm of CPS's first four generations --- where battles are longest and reveal the least information about the opponent's team --- our largest RL policy is officially estimated to have a $41$-$58\%$ chance to defeat a randomly sampled opponent (depending on the generation). Rather than waiting for more data to accumulate in our dataset, we explore the idea that our models would benefit from training on intentionally unrealistic self-play data that does not attempt to recreate the unknown distribution of teams and opponents in online battles. The resulting agents improve to win rates of $64$-$80\%$ --- rising into the top $10\%$ of active usernames and onto the global leaderboards. A recent LLM Agent \citep{hu2024pokellmon} proves uncompetitive in the long horizons of the early generations, and our best agents match or surpass the strongest heuristic search engine.

\vspace{-1mm}
\section{Background: Competitive \poke Singles}
\vspace{-1mm}

If the reader is unfamiliar with Competitive \poke\hspace{-1mm}, it is difficult to overstate how complicated top-level strategy can be. The game combines opponent modeling with stochastic transitions, complex dynamics, long-horizon planning, and a large initial state space. \poke is highly stochastic, and gameplay revolves around nuanced mechanics with endless edge cases. CPS is played on \href{https://pokemonshowdown.com/}{\poke Showdown} (\textbf{PS}) --- a website with thousands of daily players. PS simulates the combat mechanics of each major commercial game release (or ``generation''). Some fundamentals transfer, but competitive play relies on details specific to each generation. PS divides generations into ``tiers'' that enforce various rules to maintain competitive balance. Each tier of each generation is its own game --- or rather, two games played consecutively: team \textit{design} and \textit{control}. Players design teams before they are matched against an opponent and make trade-offs to counter threats they believe they may face. Team design converges to an equilibrium that helps narrow the search to perhaps many thousands of meaningfully distinct teams that are considered competitively viable.

In addition to navigating \poke\hspace{-1mm}'s randomness, team control (battling) focuses on decision-making under imperfect information. Details of the opponent's \poke\hspace{-1mm} are only revealed when they directly impact the battle. We can gain an advantage by inferring our opponent's team based on what they have already revealed. For example, we might know that \poke $A$ is often used alongside \poke $B$ and that \poke $A$ commonly brings move $x$ or $y$ but rarely brings both. We may try to mislead our opponent by revealing information that suggests one team design only to surprise them later in the battle. Players make (most) decisions simultaneously. Accurately predicting the opponent's choices based on their team and previous tendencies is the key skill that differentiates high-level players. For example, a move may win the battle but only be safe to select if we believe our opponent will switch their \poke on this turn. In short, \poke players are constantly updating a prior over the opponent's team and strategy to improve their decision-making.

\begin{figure}[h!]
    \centering
    \includegraphics[width=.8\linewidth]{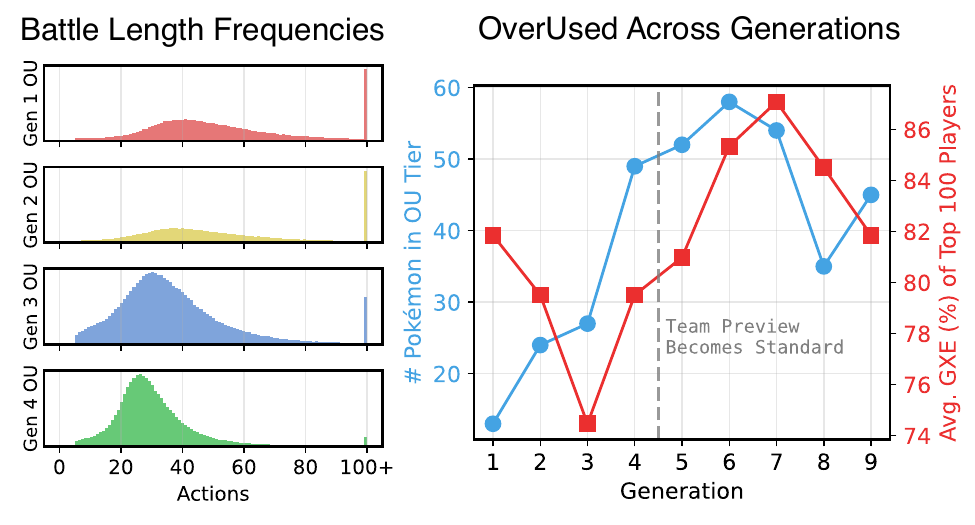}
    \vspace{-5mm}
    \caption{\textbf{Episode Length, Team Diversity, and Variance by Gen.} Battle lengths are based on our replay dataset and binned with a max length of $100$. GXE statistics are captured in February 2025.}
    \label{fig:ps_gens_summarized}
    \vspace{-2mm}
\end{figure}

There are three player metrics on PS. \textbf{ELO} is a standard rating system, but PS's version is intentionally noisy, and ELO is not comparable across game modes. \textbf{Glicko-1} is an ELO-like rating that considers the full history of a player's battles and is a much better estimate of true skill for our purposes. The matchmaking system on PS prefers to pair players with similar ELO ratings. \textbf{GXE} corrects for this matchmaking bias to estimate a player's odds of defeating a randomly sampled opponent. \poke has the kind of inherent variance that would be familiar to Heads-Up No-Limit Texas Hold'em players: minimizing risk is considered a key skill, but some losses are inevitable. The very best players have a GXE between $74$-$90\%$ (Figure \ref{fig:ps_gens_summarized} Right).

\vspace{-3mm}

AI research in PS faces the question of which generation and tiers to study. The standard choice is the most recent generation's ``random battles'' tier. Random battles remove team design by providing each player with a procedurally generated team. This ruleset has a more casual player base, and we will focus on formats where players design teams tailored to their playstyle. \textbf{Our agents will learn to play four different tiers, but evaluations will focus on ``OverUsed'' (OU)}. OU is the definitive competitive format, making it the most popular and, therefore, the tier with the most data to learn from (Section \ref{sec:dataset}). Broadly speaking, each generation of OU increases the number of team combinations and gameplay mechanics (Figure \ref{fig:ps_gens_summarized} Right). Importantly, the size of the team space creates so much variance from Generation 5 onwards that PS adopts a mechanic called ``team preview'' that reveals the opponent's team before the start of the battle. We are particularly interested in the partial observability of CPS. For this reason, we focus on the first four generations.

\textbf{Early Generation OverUsed.} In addition to their signature lack of team preview, the early generations of CPS are defined by their unique gameplay mechanics and outlier battle lengths (Fig. \ref{fig:ps_gens_summarized} Left). Gen1 and Gen2 are infamously stochastic, and reduced offensive power shifts focus away from team composition and towards battle strategy over long exchanges. Gen3 is notable for its enduring popularity and competitive balance --- with a narrow margin between median and top-level players by GXE (Fig. \ref{fig:ps_gens_summarized} Right). Gen4 resembles modern versions in that many \poke can eliminate their opponent in a single move; the fast pace of play leads to high-stakes decisions over short planning depths. The early generations are an almost independent competitive community with a long history and a relatively small but self-selective player base. The people we will be playing against have intentionally sought out the competitive format of a $15+$ year-old game because it is their interest and expertise. There are few casual players here; many of the ``low rated'' usernames we will face are experienced players logged into alternate accounts for various reasons. Appendix \ref{app:heuristics} finds that a heuristic using basic \poke principles and lookup tables is far less effective against human players in early-generation OU than modern random battles.

While our use of model-free long-context RL and focus on Early Gen OU are novel, there is existing work on AI for CPS. The best \poke bots focus on heuristic tree search with custom high-throughput simulators. Some work has experimented with network-based state evaluation and Monte Carlo tree search (MCTS) \citep{mcts_survey} for random battles formats \citep{ASelfPlayPolicy}. \poke is primarily played and discussed on the internet, and this affords considerable gameplay knowledge to recent LLM-Agent techniques \citep{hu2024pokellmon, karten2025pokechamp}. Key baselines will be discussed as we play against them in Section \ref{sec:experiments}. Appendix \ref{app:pokemon_related_work} provides a survey of AI in CPS, while Appendix \ref{app:general_related_work} discusses related work in offline RL and gameplaying.

\vspace{-2mm}

\section{Building an Offline RL Dataset of Real Human Battles}
\label{sec:dataset}

PS creates a log (``replay'') of every battle that expires after a brief period unless saved. Players save replays for later study, to share a fun outcome with friends, or as a way to record official tournament results. PS has been the home of Competitive \poke for over a decade --- time enough to accumulate millions of replays. The PS replay dataset is an exciting source of naturally occurring data. However, there is a critical problem: CPS decisions are made from the partially observed point-of-view of one of the two competing players, but PS replays record the perspective of a third-party spectator who has access to information about neither team. We unlock the PS replay dataset by converting spectator views to each player's perspective separately.

Replay reconstruction involves four high-level steps. First, we \textbf{simulate} the current state of the battle from a spectator perspective according to the PS API. Throughout this process, we use incoming information to estimate the initial configuration of both unobserved teams. At the end of the battle, we \textbf{infer} any information that was never revealed. To do this, we need a way to model the distribution of competitive teams in each generation and tier. Fortunately, the PS community tracks \poke usage statistics to measure trends and evaluate rule changes. We use available usage data and the revealed teams of similar replays to model the distribution of human-constructed teams. Next, we \textbf{backfill} inferred team rosters for a chosen point-of-view player to replicate the information they would have observed when their decisions were made. Finally, we \textbf{convert} the reconstructed trajectory to a format identical to the online simulator. Appendix \ref{app:replay_reconstruction} walks through a simplified example and uses a real replay to visualize the raw input, inferred team, and trajectory output according to the observation space, action space, and reward function discussed in the next section.

\begin{figure}[h!]
    \centering
    \vspace{-4mm}
    \includegraphics[width=.9\textwidth]{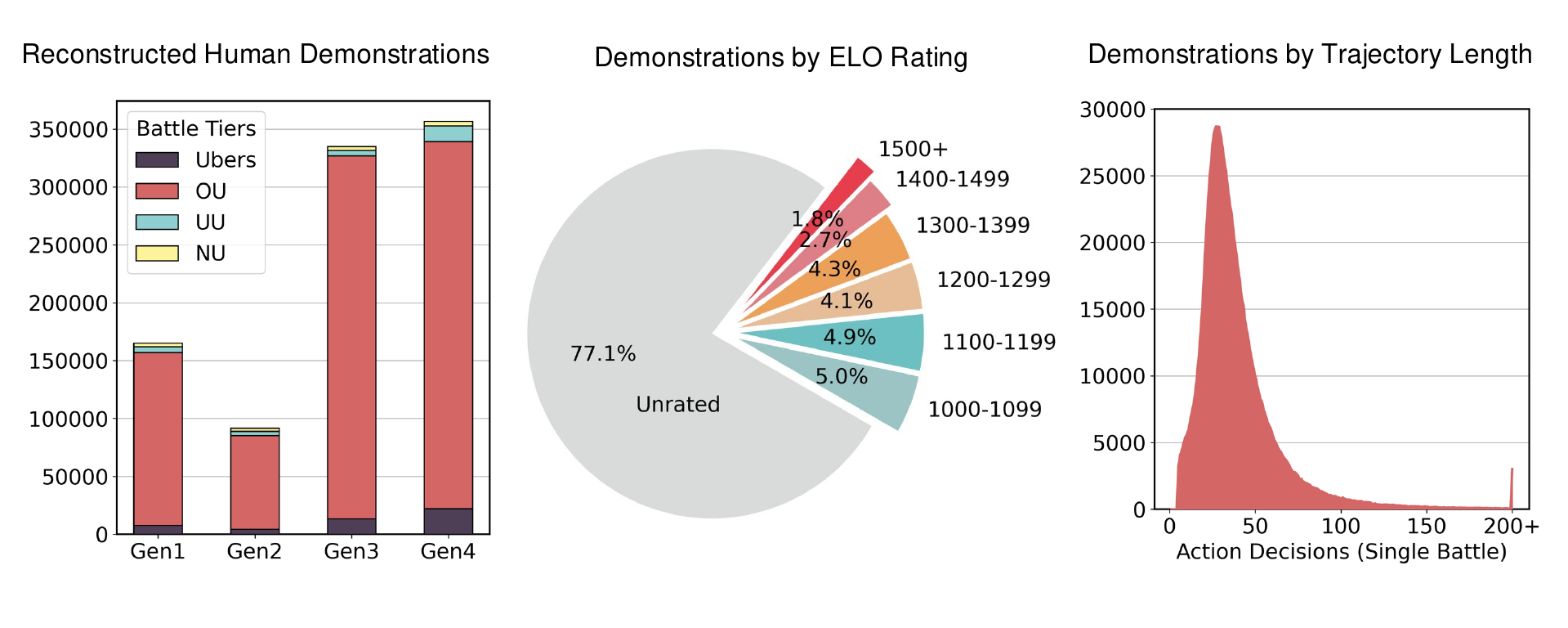}
    \vspace{-7mm}
    \caption{\textbf{Dataset Summary.} The initial version of our offline dataset includes $475$k battles --- summarized here by their PS format (left), ELO rating (center), and length in agent timesteps (right).}
    \label{fig:offline_dataset}
\end{figure}

This process is not always successful, as some gameplay mechanics cannot be reconstructed from incomplete information. A list of checks identifies trajectories that have entered ambiguous situations and conservatively discards them. All told, we are able to download and reconstruct more than $475$k human demonstrations (with shaped rewards) from historical Gen $1$-$4$ battles dating back to $2014$ (Figure \ref{fig:offline_dataset}). Each battle yields two point-of-view trajectories for a total of about $950$k sequences containing $38$M timesteps. Player names and chats are anonymized, and trajectories are stored in a flexible format that lets researchers customize observations, actions, and rewards. Our pipeline is actively downloading new battles and has recently expanded to include Gen 9 OU, bringing the total to $3.5$M trajectories. However, the experiments in this paper use the original $950$k-trajectory dataset, with a cutoff of September 2024.

\vspace{-1mm}

\section{Search-Free \poke with Offline RL On Sequence Data}
\label{sec:method}

\vspace{-1mm}

Players discuss and teach the game based on the idea that their decision-making policy $\pi$ is conditioned on their current estimate of their opponent's policy ($\pi_o$) and team \textit{c}omposition ($c_o$). Let $c_p$ be our own team composition. This paper will take a Bayesian RL \citep{NIPS2007_3b3dbaf6, ghavamzadeh2015bayesian} or meta-RL \citep{beck2023survey} perspective where we consider our opponent's choices part of the environment's unknown transition function $T(s_{t+1} \mid s_t, a_t, \pi_o)$ \citep{zintgraf2021deep}. Our goal is to find a policy that maximizes return over some distribution of latent environment variables, which in our case would be the opponents active on PS and our distribution of teams:

\vspace{-8mm}

\begin{align}
\pi^* = \arg\max_{\pi} \mathbb{E}_{\pi_o,\, c_o \sim p(\pi_o, c_o), c_p \sim p(c_p)} \left[
\mathbb{E}_{\tau \sim p(\tau \mid \pi, \pi_o, c_o, c_p)} \left[ \sum_{t=0}^{T} \gamma^t\, R(s_t, a_t) \right]
\right]
\label{eq:objective}
\end{align}

\vspace{-3mm}

Context-based methods condition the policy on estimates of the unobserved variables derived from previous experience. Here, this would amount to using the entire history of a battle\footnote{A natural extension of the context-based framework here would include previous battles between the same players alongside their current battle. This may allow for adaptation in a tournament best-of-three match format.} (observations, rewards\footnote{Because our \poke reward function never changes, it would be considered part of the state space and happens to be important for inferring the \textit{outcome} of the previous turn in our setup.}, and the actions of both players) to estimate ($c_o$, $\pi_o$). If we want to avoid explicitly predicting $c_o$ or $\pi_o$ \citep{humplik2019meta} (which is difficult to formulate) or modeling the complicated dynamics of \poke \citep{zintgraf2021varibad}, we can follow a simple black-box framework \citep{duan2016rl, wang2016learning} where a sequence model $S_{\theta}$ takes all prior experience under the current latent variables (the entire battle up until the current timestep, $\tau_{0:t}$) as input and outputs a representation $h_t$ for the policy network $\pi_{\phi}$. The system is trained end-to-end to maximize Eq. \eqref{eq:objective} as in standard deep RL. Because a better estimate of the opponent will increase win rate, the sequence model will implicitly learn that behavior. The policy navigates an exploration-exploitation trade-off at test time, where it may take actions that reveal new information if this increases expected returns.

We will be using the offline dataset ($\mathcal{D}$) from Section \ref{sec:dataset} to approximate the expectations in Eq. \eqref{eq:objective}, which assumes that the distribution of teams and playstyles across history is identical to that of the current game \citep{dorfman2020offline, NEURIPS2024_8a30aba6}. This is false, but it may be close enough, particularly in the optimized world of Early Gen OU. If we want to expand our dataset (i.e., by self-play), we need to try to select teams and opponents that match the true distribution. Alternatively, we can collect data that is \textit{unambiguously} out-of-distribution (OOD). For example, we can place a rare \poke in the lead-off position so that when the policy begins a real battle and sees a more standard choice, it has no reason to believe it is facing our synthetically generated teams or opponents. 

\poke has a complex state space, and our policy may need to be large and non-trivial to train with offline RL. To stabilize, we can frame the problem from a behavior cloning (BC) perspective: predicting the actions of a human player requires reasoning about the strategy of the player we are imitating and their understanding of the opponent. Accurate predictions will require long context inputs. RL is a tool to sort through the noise of a large dataset that includes the decisions of all levels of players in both competitive and casual settings. We arrive at the same setup but prefer an update that safely reduces to BC while allowing room to skew the loss function towards return-maximizing behavior if we decide the offline RL risks are sufficiently small \citep{springenberg2024offline, wu2019behavior, fujimoto2021minimalist}. Ideally, BC becomes a lower bound upon which we can improve. Solutions of this kind are actor-critics that train their critic to output $Q$-values with standard one-step temporal difference backups. Actor loss functions take the general form:
\vspace{-8mm}

\begin{align}
    \mathcal{L}_{\text{Actor}} = \mathbb{E}_{\tau \sim \mathcal{D}}\left[\frac{1}{T}\sum_{t=0}^{T} \left(-w(h_t, a_t)\log \pi(a_t \mid h_t) - \lambda \mathbb{E}_{a \sim \pi(\cdot \mid h_t)}\left[Q\left(h_t, a\right)\right]\right)\right]
    \label{eq:actor_objective}
\end{align}

\vspace{-4mm}

\begin{wraptable}{r}{.42\textwidth}
\vspace{-3mm}
\centering
\resizebox{.40\textwidth}{!}{%
\begin{tabular}{@{}ccc@{}}
\toprule
\textbf{\begin{tabular}[c]{@{}c@{}}Model Name\end{tabular}} & \textbf{$w(h, a) = $}                                                           & \textbf{$\lambda$} \\ \midrule
"IL"                                                                & 1                                                                               & 0                     \\ \midrule
\begin{tabular}[c]{@{}c@{}}"Exp"\\ (or just "RL")\end{tabular}      & \begin{tabular}[c]{@{}c@{}}$\exp(\beta A^{\pi}(h, a)) $\\ (clipped)\end{tabular} & 0                     \\ \midrule
\begin{tabular}[c]{@{}c@{}}"Binary"\\ \end{tabular}       & $A^{\pi}(h, a) > 0$                                                             & 0                     \\ \midrule
\begin{tabular}[c]{@{}c@{}}"Binary+MaxQ''\end{tabular}      & $A^{\pi}(h, a) > 0$                                                             & > 0                   \\ \bottomrule
\end{tabular}

}
\caption{\textbf{$\mathcal{L}_{\text{actor}}$ Configurations (Eq. \eqref{eq:actor_objective})}. Advantages are estimated by the critic: $A^\pi(h, a) = Q(h, a) - \mathbb{E}_{a' \sim \pi}[Q(h, a')]$.}
\label{tbl:rl_configs}
\vspace{-1mm}
\end{wraptable}
Where $h_t$ is the output of the sequence model $S_{\theta}(\tau_{0:t})$ that replaces the state. The first term is a BC objective that re-weights decisions according to a function $w$ and constrains learning to actions taken in the offline dataset \citep{wang2020critic, nair2020awac}. The second term is the standard online off-policy actor update that risks overestimating the value of OOD actions when used offline \citep{kumar2019stabilizing}. Our experiments will study configurations of Equation \eqref{eq:actor_objective} summarized by Table \ref{tbl:rl_configs}. For further discussion of RL engineering details, we refer the reader to the AMAGO \citep{grigsby2024amago} implementation used throughout our experiments. 
\begin{wrapfigure}{r}{.52\textwidth}
    \vspace{-3mm}
    \begin{center}
    \includegraphics[width=\linewidth]{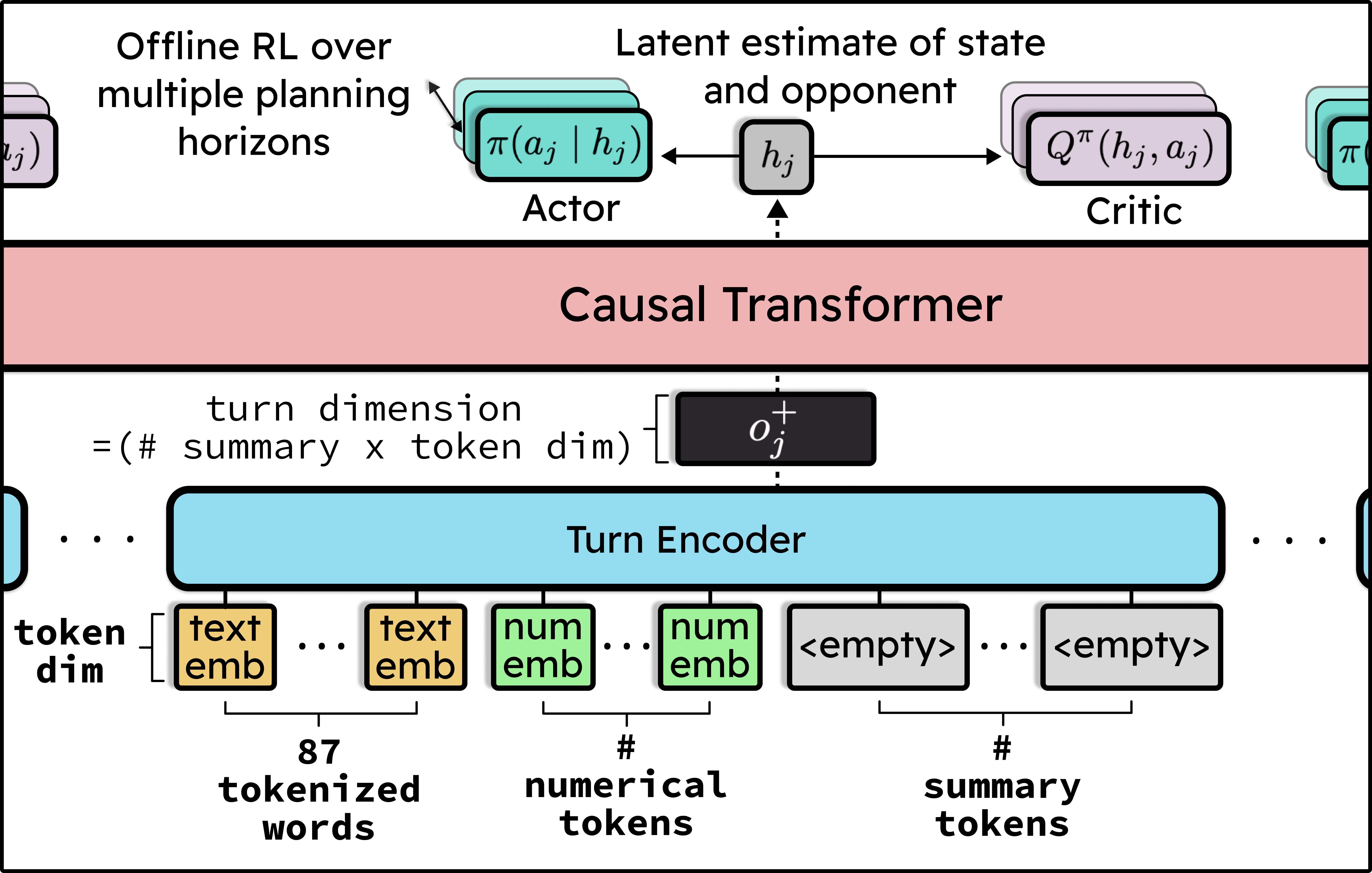}
    \vspace{-5mm}
    \caption{\textbf{Model Overview.} Actions are predicted based on representations of the observation, action, and reward of each turn in the current battle.}
    \label{fig:architecture}
    \end{center}
    \vspace{-4mm}
\end{wrapfigure}

\begin{figure}[h!]
    \centering
    \vspace{-6mm}
    \includegraphics[width=1.0\linewidth]{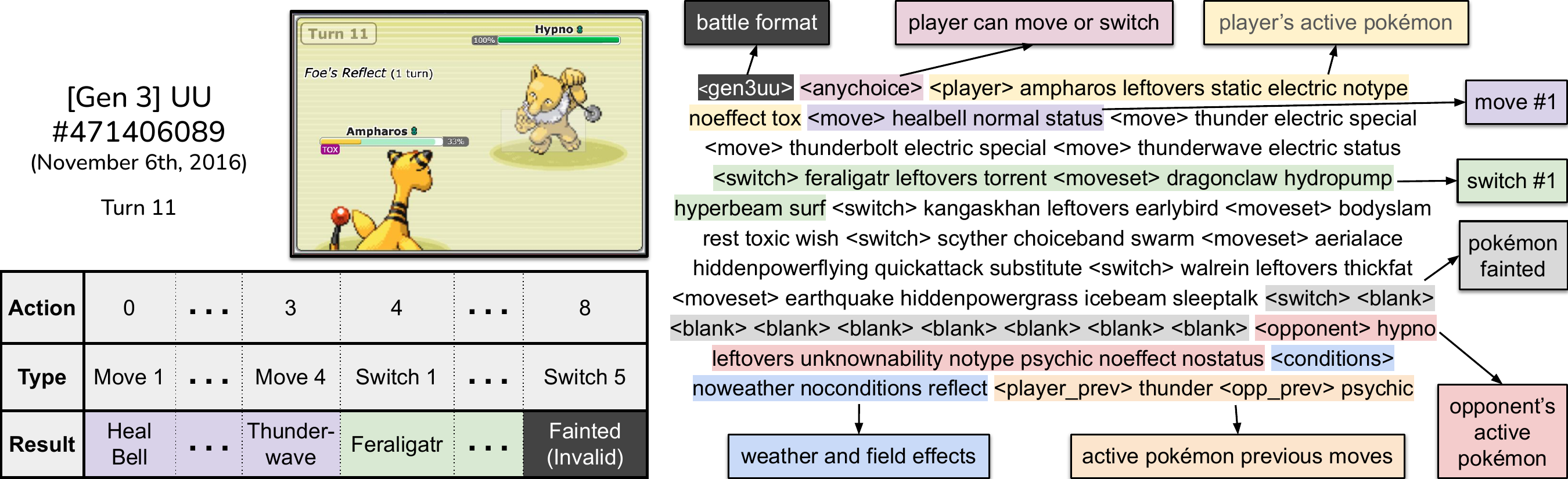}
    \vspace{-4mm}
    \caption{\textbf{Observation and Action Space.} Text order is important, but words can be tokenized into arrays with a consistent length (of $87$). Observations also include $48$ numerical features. The meaning of each action index varies by turn but is presented in the text in a consistent order.\vspace{-2mm}}
    \label{fig:annotated_obs}
\end{figure}

Next, we need to define an observation space, action space, and reward function for CPS. Our agent needs enough information to mirror human decisions, and the user interface of the PS website is an obvious point of reference. However, our models have memory, and we do not need to provide all of this information at every timestep. We have a trade-off between dimensionality, memory difficulty, generalization over \poke\hspace{-1mm}'s complex dynamics, and exposure to sim2real errors between replay reconstruction and deployment. We settle on a compromise of $87$ words of text and $48$ numerical features. The text component is semi-readable, and Figure \ref{fig:annotated_obs} provides an example from a replay in our dataset. \textbf{The most important detail is that we are relying entirely on memory to infer the opponent's team}; observations only include the opponent's active \poke\hspace{-1mm}. The memory demands of our CPS observations are more comparable to those in the commercial video games than the PS web interface. We are confident in our sequence models' ability to recall previous timesteps, and this makes it worth avoiding distribution shift over features of the opponent's full team as it is slowly revealed. There are nine discrete actions, where the first four indices correspond to the moves of the active \poke\hspace{-1mm}, and the remaining five switch to another team member. The observation conveys the precise meaning of these actions in a predictable order. The reward function is dominated by binary win/loss but includes light shaping for damage dealt and health recovered. Appendix \ref{app:training_details} provides more details.

The observation, previous action, and previous reward at each timestep are processed by a Transformer encoder that uses designated summary tokens to attend over the multi-modal sequence \citep{devlin2019bert}. Text is encoded by tokenizing the \poke vocabulary based on our dataset with an \texttt{<unknown>} token for rare cases we may have missed\footnote{We experiment with an augmentation scheme that sets tokens to \texttt{<unknown>} to force recovery from previous timesteps. Models above $100$M parameters use this strategy by default, while its use in smaller models is indicated by ``Aug.'' We do not find evidence that this strategy impacts performance.}. The resulting sequence of turn representations is the input to a causal Transformer with actor and critic output heads (Figure \ref{fig:architecture}).

\section{Experiments}
\label{sec:experiments}

We will begin evaluating a progression of increasingly RL-heavy training objectives across model architectures with  ``Small'' (15M), ``Medium'' (50M), and ``Large'' (200M) parameter counts summarized by Table \ref{app:hparams}. Models are named in results according to their size and training objective (Table \ref{tbl:rl_configs}). Table \ref{tbl:models} provides a complete list of model configurations. Results will be discussed in semi-chronological order, though some figures will spoil win rates of models trained on ``synthetic'' self-play datasets described in Section \ref{sec:self_play}. Our goal is to compete against human players, but this is expensive and creates a challenging evaluation problem: Which model checkpoints do we deploy on PS? Our efforts to answer this question result in extensive evaluations against various opponents.

Training uses the offline dataset to assign our players' teams, but we need to ``prompt'' our agents with a set of teams during evaluations. We use three sets: 1) The \textbf{Variety Set} procedurally generates $1$k intentionally diverse teams per gen/tier and will be used to evaluate OOD gameplay and to generate unambiguous self-play data as mentioned in Section \ref{sec:method}. 2) The \textbf{Replay Set} approximates the choices of top players based on their replays and infers unrevealed details as done in Section \ref{sec:dataset}. 3) The \textbf{Competitive Set} comprises $10$-$20$ complete ``sample'' teams per gen/tier scraped from forum discussions; these are generally designed for beginners by experts. Win rates are measured over large samples of hundreds or thousands of battles unless otherwise noted. Evaluations use \href{https://github.com/hsahovic/poke-env}{\texttt{poke-env}} \citep{poke-env} to interact with a locally hosted PS server and the public website.

\vspace{-2mm}

\subsection{Heuristic Evaluations}

\label{sec:experiments:heuristics}

\begin{figure}[h!]
    \centering
    \vspace{-5mm}
    \includegraphics[width=0.85\linewidth]{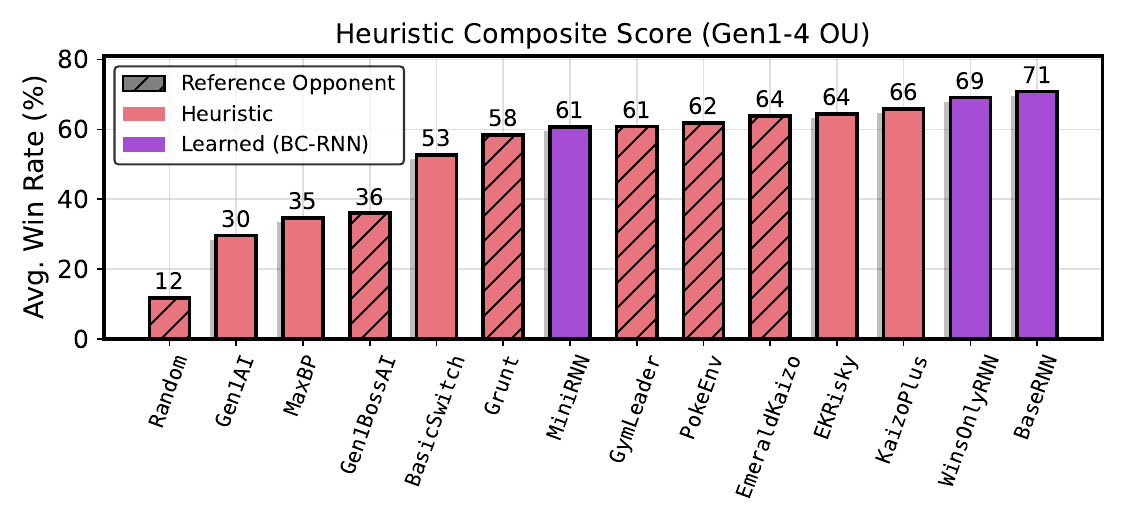}
    \vspace{-5mm}
    \caption{\textbf{Heuristic Composite Scores.} The average win rate against six of our heuristics measures core game knowledge and creates a relatively fixed point of reference across different game modes.}
    \vspace{-3mm}
    \label{fig:heuristic_composites}
\end{figure}

\begin{wrapfigure}{r}{.52\textwidth}
\begin{center}
    \centering
    \vspace{-7mm}
    \includegraphics[width=\linewidth]{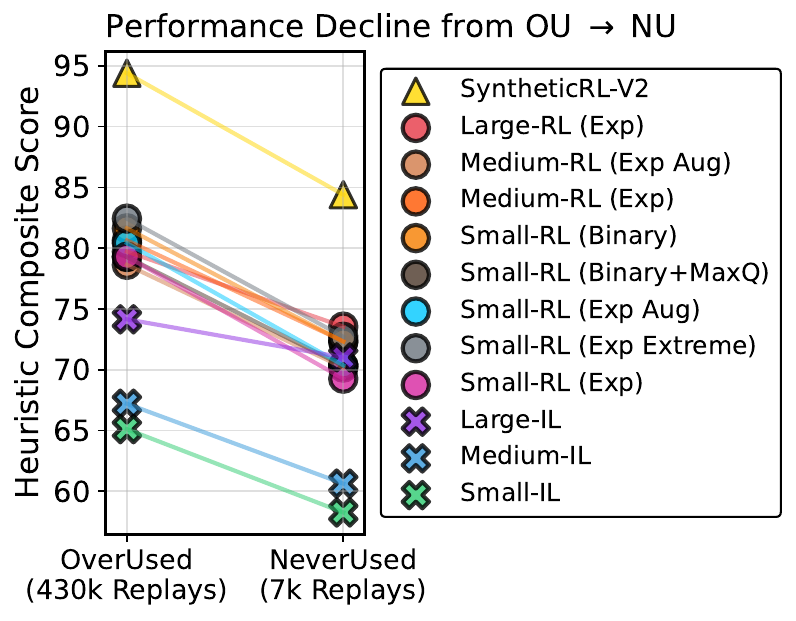}
    \vspace{-8mm}
    \caption{\textbf{OU $\rightarrow$ NU.} Heuristics highlight a gap between OU tiers and those with fewer replays. OU scores are directly comparable against Fig. \ref{fig:heuristic_composites}.}
    \label{fig:ou_vs_nu}
    \vspace{-6mm}
\end{center}
\end{wrapfigure}

We create a suite of a dozen heuristic opponents that evaluate core game knowledge. Strategies are based on fundamental \poke concepts and re-implementations of policies from official versions of \poke\hspace{-1mm}, fan-made ROM hacks with inflated difficulty, and popular CPS AI baselines. Full descriptions of these policies and their relative performance are provided in Appendix \ref{app:heuristics}. The average win rate against $6$ of these heuristics on the Variety Set forms a ``Heuristic Composite Score'' (Figure \ref{fig:heuristic_composites}). We tune the Turn Encoder architecture (Fig. \ref{fig:architecture}) with RNN trajectory models $S_{\theta}$ between $500$k-$4$M parameters trained by BC. Appendix \ref{app:base_rnn} documents the predictive accuracy of these models and provides further details. The best BC-RNN models lead the early Heuristic Composite rankings, and these will become the next rung on the ladder toward human-level gameplay. Clear signs of underfitting motivate the starting point of $15$M for our Transformer agents. While we will go on to saturate this benchmark in OU, heuristics represent a fixed target unaffected by the discrepancies in data availability between OU and the other three tiers our agents are trained to play (Fig. \ref{fig:offline_dataset}). Figure \ref{fig:ou_vs_nu} documents a predictable decline from OU to NeverUsed (NU) gameplay. We evaluate many variants of the $\mathcal{L}_{\text{actor}}$ objective (Eq. \ref{eq:actor_objective}) but do not find significant differences between them.
\vspace{-6mm}

\subsection{Model-Based Evaluations}
\vspace{-1mm}

\begin{wrapfigure}{r}{.60\textwidth}
    \begin{center}
    \vspace{-20mm}
    \includegraphics[width=\linewidth]{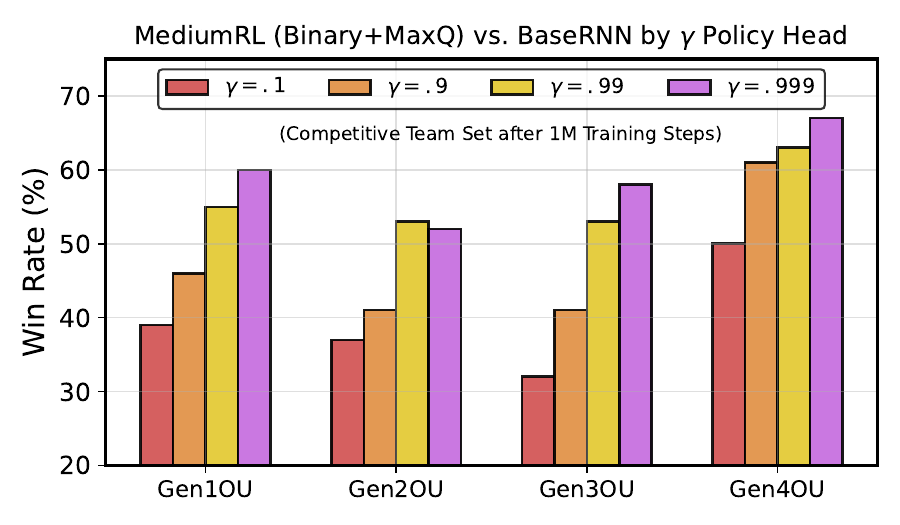}
    \vspace{-8mm}
    \caption{\textbf{Multi-$\gamma$ Policies.} Models train over multiple value horizons, but long-term planning increases win rate.}
    \label{fig:multigamma_eval}
    \end{center}
    \vspace{-4mm}
\end{wrapfigure}
Appendix \ref{app:base_rnn} evaluates our larger Transformer models against our best RNN baseline. RL updates significantly outperform the pure-BC Transformers, but there is little difference between the many RL variants considered. The expected relationship between model size and performance is clearer for BC than it is for RL. Following \citet{grigsby2024amago}, we are optimizing actor and critic network outputs for a set of $\gamma$s in parallel. At test time, we are able to select the action corresponding to any of these horizons. Figure \ref{fig:multigamma_eval} verifies that our agents are using long-term value estimates to improve their win rate. All other evaluations follow the policy for $\gamma=.999$. With RL comfortably outplaying our smaller IL baselines on the more limited Competitive Team Set, we shift to playing against Large-IL on the Replay Set. Figure \ref{fig:vs_large_il} highlights the win rate of key models in OU.

\subsection{Synthetic Data from Self-Play}
\label{sec:self_play}
Section \ref{sec:exp:humans} will find that our offline dataset yields policies capable of human-level gameplay on the public ladder. Our agents contribute to each day's batch of new replays and grow the dataset alongside human players. In principle, we could wait to retrain new policies on a larger dataset, but this data is not making a significant difference on the timescale of a single project. We can speed up the process by deploying agents on a local PS ladder, adding their trajectories to the human gameplay dataset, and retraining or fine-tuning (Figure \ref{fig:fig1} Left). However, we need to be wary of a shift between the frequency of teams and opponents implied by the new offline dataset and the true distribution on PS. One approach would be to try and generate data that is clearly different from the original set so that when conditioned on a real battle, our model's implicit estimate of $p(\pi_o, c_o \mid \tau_{0:i})$ should be unchanged at small $i$. We let a mix of checkpoints from all our agents compete on a locally hosted PS ladder, playing with teams from the Variety Team Set. By prioritizing diversity over realism, we hope this data will cover replay reconstruction failures and improve model-free learning of Pokémon's stochastic transitions without biasing estimates of human teams and strategies.

\begin{figure}[h!]
    \centering
    \vspace{-8mm}
    \includegraphics[width=\linewidth]{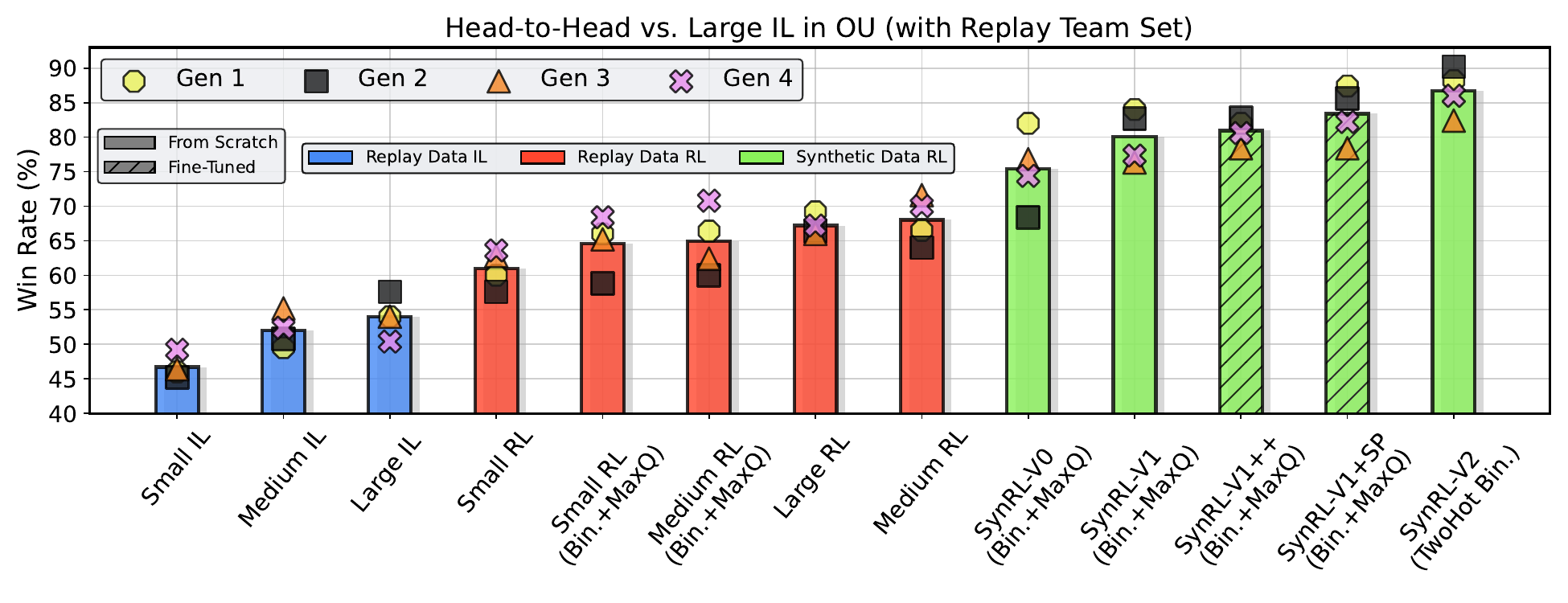}
    \vspace{-8mm}
    \caption{\textbf{Self-Evaluation Against Large IL.} Results are determined by the best checkpoint over the last $200$k training steps with a sample size of $500$ battles per generation.}
    \vspace{-3mm}
    \label{fig:vs_large_il}
\end{figure}

The SyntheticRL (SynRL) models are Large Binary+MaxQ (Eq. \eqref{eq:actor_objective}) policies trained from scratch. \textbf{SyntheticRL-V0} trains on ``synthetic'' variety data for generations $1$ and $3$ only, for a total dataset size of $2$M trajectories. It is a promising improvement over our previous policies against heuristics (Fig. \ref{fig:heuristic_learning_curves}), BC-RNN (with win rates as high as $95\%$ in Gen1OU and $85\%$ in Gen3OU), and Large-IL (Fig. \ref{fig:vs_large_il}). \textbf{SynRL-V1} takes this dataset and adds generations $2$ and $4$ to reach a total of $3$M trajectories (retraining a $200$M policy from scratch) for a consistent improvement across generations.
\begin{wrapfigure}{r}{.42\textwidth}
    \begin{center}
     \vspace{-5mm}
    \includegraphics[width=\linewidth]{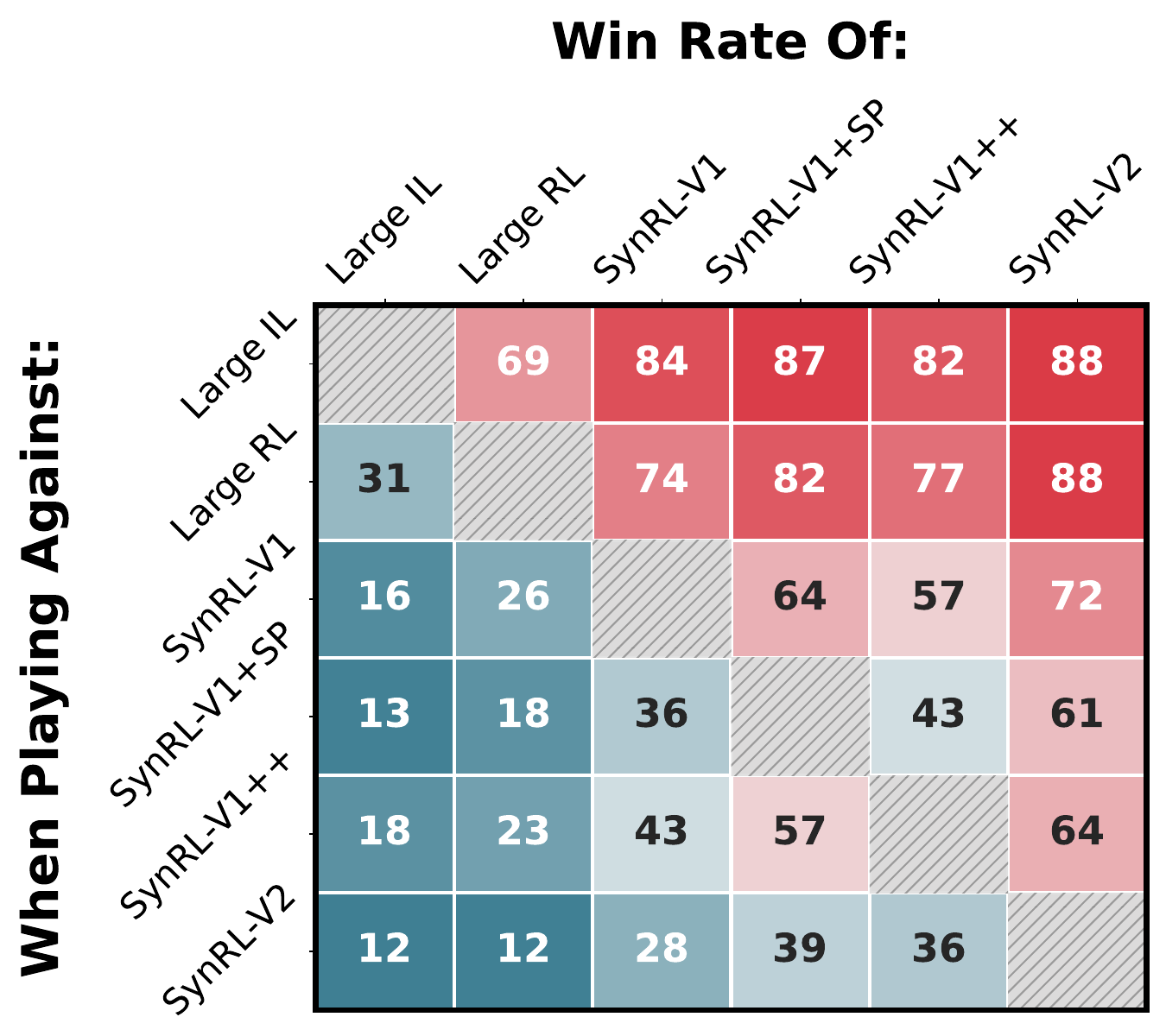}
    \vspace{-6mm}
    \caption{\textbf{Gen1OU Self-Play.} Sample of $500$ battles on the Replay Set.}
    \label{fig:self_play_win_matrix}
    \vspace{-5mm}
    \end{center}
\end{wrapfigure}

We might wonder whether the caution of the ``synthetic'' data process was necessary. We test this by letting SynRL-V1 battle recent checkpoints of itself with the more realistic Replay Set until the offline dataset is $5$M trajectories. Afterward, we resume training for another $200$k gradient steps to create \textbf{SynRL-V1+SelfPlay (SP)}. As expected, the resulting model is significantly better against itself (Figure \ref{fig:self_play_win_matrix}), and a key baseline in Section \ref{sec:exp:external_baselines}, but this will translate to inconsistent improvement against real players in Section \ref{sec:exp:humans}. Battle replays make it clear that the model believes it is playing SynRL-V1. We backtrack and expand the dataset to $5$M with unrealistic teams and IL opponents and fine-tune SynRL-V1 again to create \textbf{SynRL-V1++}. Finally, we train a new $200$M model from scratch on the SynRL-V1++ dataset with $50$k new human replays. Instead of Binary+MaxQ, we use a simple binary weighted BC update with value prediction converted to two-hot classification \citep{schrittwieser2020mastering, hafner2023mastering, farebrother2024stop} as implemented in this setting by \citet{grigsby2024amago2}. This trick is often motivated by hyperparameter insensitivity and invariance to return magnitudes in multi-task RL. In our case, improved critic accuracy leads to an entirely new level of pessimism in the binary BC filter (Figure \ref{fig:binary_filters}) --- a potential improvement considering our dataset is now primarily composed of decisions made at beginner or intermediate human levels (Sec. \ref{sec:exp:humans}). The resulting \textbf{SynRL-V2} model is our best by every metric (against heuristics, other models, and key external baselines yet to be discussed).

\subsection{LLM Agents and Heuristic Search}
\label{sec:exp:external_baselines}

\begin{figure}[h!]
    \centering
    \begin{subfigure}{0.56\linewidth}
        \centering
       \includegraphics[width=\linewidth]{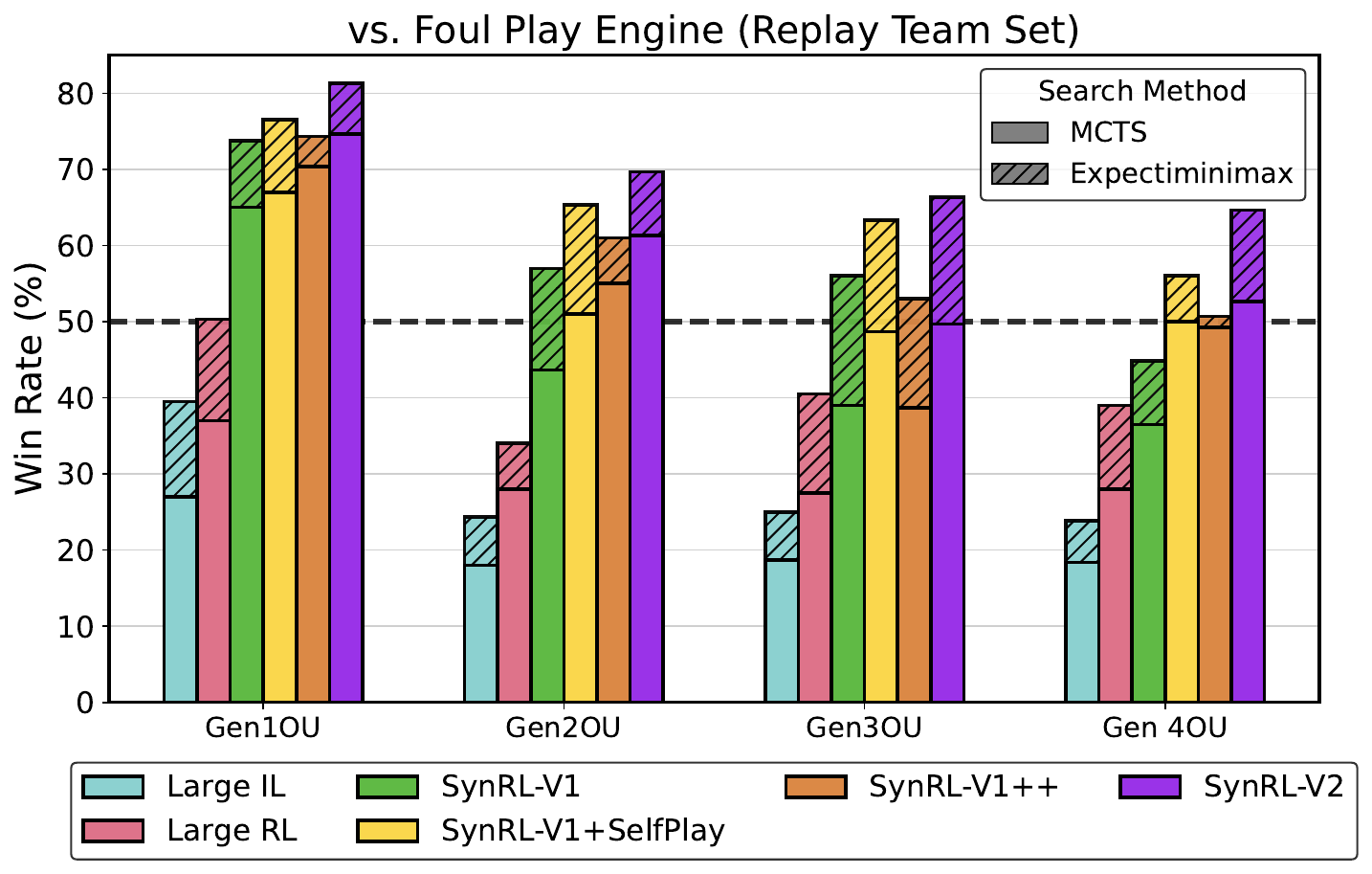}
        \vspace{-5mm}
        \caption{\textbf{Foul Play Evaluation.} Using both available search algorithms and \texttt{poke-engine} v$0.31.0$. Sample of $300$ battles.}
        \label{fig:vs_foul_play}
    \end{subfigure}
    \hfill
    \begin{subfigure}{0.43\linewidth}
        \centering
        \includegraphics[width=\linewidth]{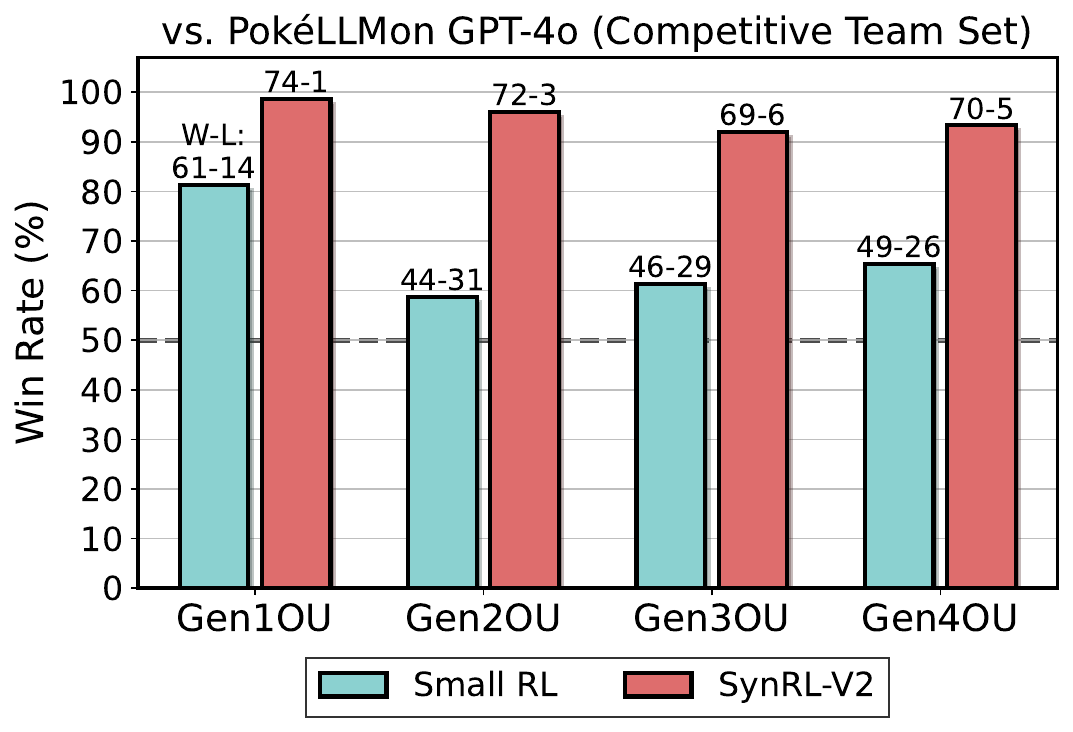}
        \vspace{-3mm}
        \caption{\textbf{PokéLLMon}. GPT-4o backend with custom prompts for Gen1-4. Sample of 75 battles.}
        \label{fig:pokellmon}
    \end{subfigure}
    \vspace{-4mm}
\end{figure}

Foul Play \citep{foul-play} is an advanced engine for CPS that uses a custom simulator to search over \poke\hspace{-1mm}'s game tree. With extensive domain knowledge, it implements much of the behavior we would hope our policies can learn from data. For example, it infers its opponent's team during battles using PS usage statistics, much like we do during dataset construction. A January 2025 update to Foul Play introduced support for the early generations. We challenge the engine to matches of $300$ battles per generation on the Replay Team Set, with results shown in Figure \ref{fig:vs_foul_play}. We manage to play the best version of the bot to a draw in Gens 3 and 4 (where the effective search depth would be lowest), and outperform it in the long horizons of Gens 1 and 2. PokéLLMon \citep{hu2024pokellmon} is a more general approach that takes advantage of \poke\hspace{-1mm}'s extensive web presence to build an LLM-Agent. Prompts are constructed with domain knowledge such as \poke type matchups and move descriptions, and the LLM is tasked with deciding between the available moves. \citet{hu2024pokellmon} evaluate in a random battles tier and note that the agent struggles with long-term planning; this effect is much more noticeable in the longer battle lengths of Gen1-4 (Figure \ref{fig:pokellmon}).

\vspace{-2mm}

\vspace{-2mm}

\subsection{Playing Humans On the \poke Showdown Ranked Ladder}

\label{sec:exp:humans}

\begin{figure}[h!]
    \centering
    \vspace{-4mm}
    \includegraphics[width=.98\linewidth]{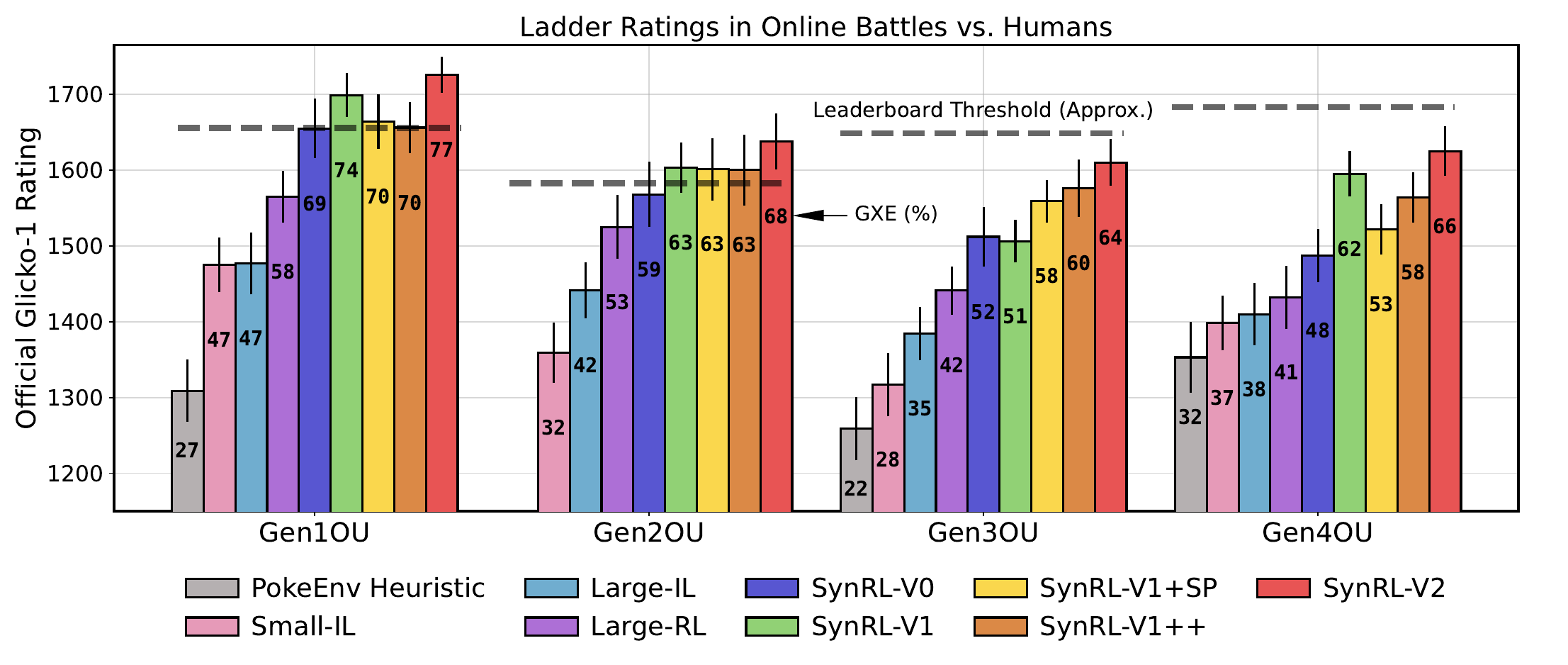}
    \vspace{-3mm}
    \caption{\textbf{Human Evaluations.} We visualize the Glicko-1 ladder rating (with its rating deviation). Bar labels represent GXE statistics. To compare across generations, we plot a heuristic baseline's performance and the average Glicko-1 of the bottom $100$ players on the Top $500$ global leaderboard.}
    \label{fig:human_evals}
\end{figure}

We compete against human players by queuing for ranked battles on the public PS ladders. We evaluate our agents over periods of $4$-$8$ days --- frequently switching between generations to sample a wider variety of opponents and achieve large sample sizes of at least $400$ battles. Evaluations run from late December 2024 through late March 2025. Models' Glicko-1 and GXE stats at the end of their final battle are shown in Figure \ref{fig:human_evals}. We include the results of a heuristic agent for additional context. Figure \ref{fig:human_percentiles} converts ladder statistics to a percentile among active usernames. Percentiles are more interpretable without a CPS background, but the distribution of player stats necessary to compute them is \textit{not} public information; PS only displays the ratings of the top $500$ active usernames. However, we are able to create a reasonable estimate because our dataset is reconstructing (most) battles played over the latter half of the evaluation period. We recover the ratings of all the unique usernames that are active enough to have a Glicko-1 deviation $\leq \pm 100$. This metric is still not ideal because players frequently use multiple usernames. Top players have clear competitive reasons to make new accounts, but we are unable to account for this. The evaluations of SynRL-V1++ and SynRL-V2 in Gen1OU are impacted by a weeks-long tournament that requires participating (top) players to make new accounts and leads to massive rating deflation in our high skill bracket\footnote{SynRL-V2 plays $613$ human battles and settles at a Gen1OU GXE of $79.9\%$ (Glicko-1 $1761 \pm 35$) after more than $100$ battles. However, its rating declines over its next $100$ games because we stop avoiding a competition where top players are playing with fresh (low-rated) usernames. Figures \ref{fig:human_evals} and \ref{fig:human_percentiles} conservatively report the final metrics.}.

\begin{wrapfigure}{r}{.35\textwidth}
    \begin{center}
    \vspace{-7mm}
    \includegraphics[width=\linewidth]{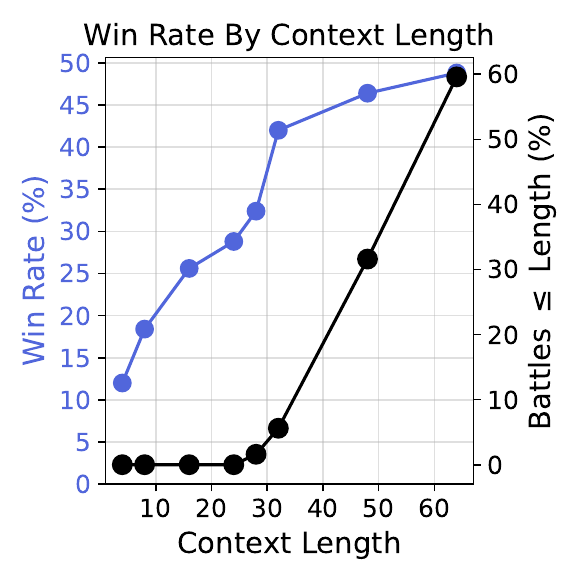}
    \vspace{-9mm}
    \caption{\textbf{Memory.} SynRL-V1 battles a version of itself that can recall the entire battle.}
    \label{fig:context_length}
    \vspace{-5mm}
    \end{center}
\end{wrapfigure}

The Large-RL model rises to the level of an intermediate player and is favored to win against a randomly selected opponent in Gens 1 and 2. High-variety self-play data leads to dramatic improvements over the course of our work. Our SynRL-V2 model is a reasonably advanced player estimated to be inside the top decile across generations. Although ELO ratings are noisy, the SynRL-V1 and SynRL-V2 models reach peak global rankings of \#$46$ and \#$31$ in Gen1OU, respectively, and SynRL-V2 makes two appearances inside the top $300$ in Gen3OU. All RL models sit inside the top $500$ in Gen2OU. To the best of our knowledge, this is the first time an AI has achieved \textit{any} of SynRL-V2's ladder ratings in \textit{any} of the Early-Gen OU tiers --- and it achieves this without a dynamics model or falling back on \poke heuristics while learning to play $16$ rulesets at the same time (Appendix \ref{app:pokemon_related_work}). Qualitatively, our models display human-like gameplay. During our evaluation process, we saved sample replays on the PS website that can be viewed by searching models' usernames (Table \ref{tab:model_usernames}) \href{https://replay.pokemonshowdown.com/}{at this link}. Policies learn to play reasonable openings, make safe \poke switches, and anticipate the moves of their opponent. However, our agents occasionally suffer from the accumulating errors we might expect from a sequence policy and can begin to make nonsensical decisions in long battles --- particularly when the opponent is playing with a rare team or uncommon strategy. Figure \ref{fig:context_length} evaluates the impact of memory on the win rate of a policy competing against the full-context-length version of itself.

\begin{figure}[h!]
    \centering
    \vspace{-3mm}
    \includegraphics[width=0.99\linewidth]{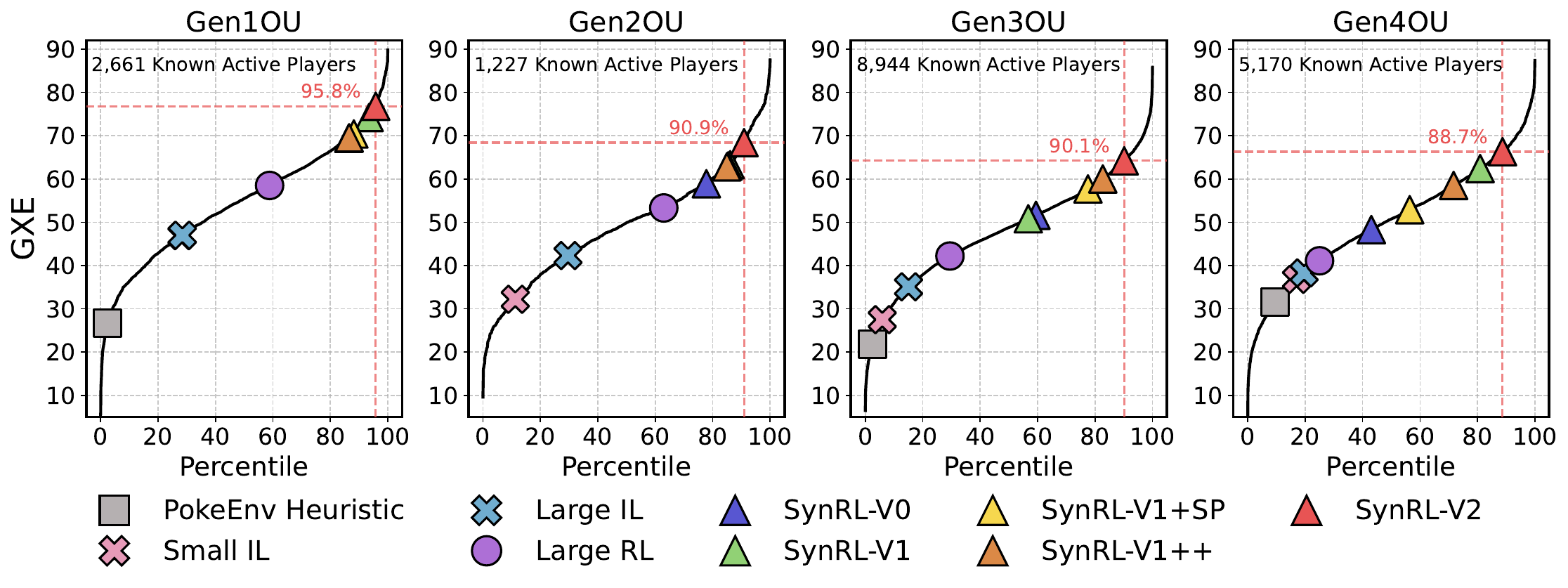}
    \vspace{-3mm}
    \caption{\textbf{Ladder Percentiles.} Replays downloaded between Feb-Mar 2025 identify $14022$ active Gen$1$-$4$ usernames. Using Gen1 as an example, $5095$ of these usernames played Gen1OU, while $2661$ were active enough to have a valid GXE statistic when results were finalized.}
    \label{fig:human_percentiles}
    \vspace{-4mm}
\end{figure}

\section{Conclusion}
\vspace{-2mm}

Our work enables a scalable offline RL approach to Competitive \poke Singles and demonstrates that sequence models trained on historical gameplay can be competitive with humans in the challenging setting of Early-Generation OverUsed. Our PS trajectory dataset will continue to grow over time and may be of broader interest in offline RL as a way to evaluate new research on a complex task. We hope our dataset and baseline models will inspire research interest in Competitive \poke\hspace{-1mm}. Alternative training details and large-scale self-play techniques may create a path to super-human performance. Our code, pretrained models, and datasets are available on GitHub at: \href{https://github.com/UT-Austin-RPL/metamon/tree/main}{\texttt{UT-Austin-RPL/metamon}}. 

% \subsubsection*{Broader Impact Statement}
% \label{sec:broaderImpact}
% In this optional section, RLJ/RLC encourages authors to discuss possible repercussions of their work, notably any potential negative impact that a user of this research should be aware of. 

\subsection*{Acknowledgments} We would like to give special thanks to Felix You and Emil Velasquez --- undergraduates at UT Austin and key early contributors to what would go on to be an unusually long research effort. Thanks also to the \texttt{poke-env} and \poke Showdown projects, as well as \poke communities like Bulbagarden and Smogon. This project would not have been possible without fan-made resources. Our research was supported by an Institute of Information \& Communications Technology Planning \& Evaluation (IITP) grant funded by the Korean Government (MSIT) (No. RS-2024-00457882, National AI Research Lab Project), a Sony Research Award, and JP Morgan.

%%%%%%%%%%%%%%%%%%%%%%%%%%%%%%%%%%%%%%%%%%%%%%%%%%%%%%%%%%%%%%%%
%% Bibliography
%%%%%%%%%%%%%%%%%%%%%%%%%%%%%%%%%%%%%%%%%%%%%%%%%%%%%%%%%%%%%%%%
\bibliography{main}
\bibliographystyle{rlj}
%%%%%%%%%%%%%%%%%%%%%%%%%%%%%%%%%%%%%%%%%%%%%%%%%%%%%%%%%%%%%%%%
%% Appendices
%%%%%%%%%%%%%%%%%%%%%%%%%%%%%%%%%%%%%%%%%%%%%%%%%%%%%%%%%%%%%%%%
\appendix

\newpage

\section{AI in Competitive \poke}
\label{app:pokemon_related_work}

\subsection{Online Tree Search}
Many CPS AI approaches rely on model-based online tree search with heuristic value approximations --- much like the methods that led to early successes in games like chess and Go. \citet{Percymon} use shallow search and mostly ignore imperfect information to reach 55$\%$ GXE in Gen6RandomBattles. The best heuristic \poke engines use PS team composition statistics to estimate private information at the current root node and reduce CPS to perfect-information depth-limited search \citep{TechnicalMachine}. \citet{TeamworkUnderExtremeUncertainty} adds more complex heuristic value functions, search pruning, and private information inference to peak at rank \#$33$ in Gen7RandomBattles. \citet{TeamworkUnderExtremeUncertainty} play a comparable number of human battles as each of our main models ($600+$). However, they are evaluating a single policy in a single ruleset --- enabling a large effective sample size that clearly demonstrates the extreme variance of PS's ELO and world ranking metrics. Glicko-1 and GXE are not reported but are far better metrics, and we encourage their use in future comparisons. Based on results in old forum posts, years of continued development, and our knowledge of method details and feature coverage relative to competitors, Foul Play \citep{foul-play} is the strongest open-source engine today.

\subsection{RL and Self-Play}

\label{app:rl_in_pokemon}

\citet{kalose2018optimal} evaluate small-scale Q-learning in a simplified version of CPS against random and minimax heuristic agents with limited success. Prior works use an online self-play process by collecting on-policy data against their own policy. \citet{ASelfPlayPolicy} train PPO \citep{schulman2017proximal} self-play agents without tree search. They achieve a 1677 Glicko-1 and 72\% GXE on the Gen7RandomBattle \poke Showdown ladder. \citet{WinningatPokémonRandomBattles} augments PPO with MCTS at test-time and achieve a 1756 Glicko-1 and 79.5\% GXE on the Gen4RandomBattle ladder.

\begin{figure}[h!]
    \centering
    \includegraphics[width=0.95\linewidth]{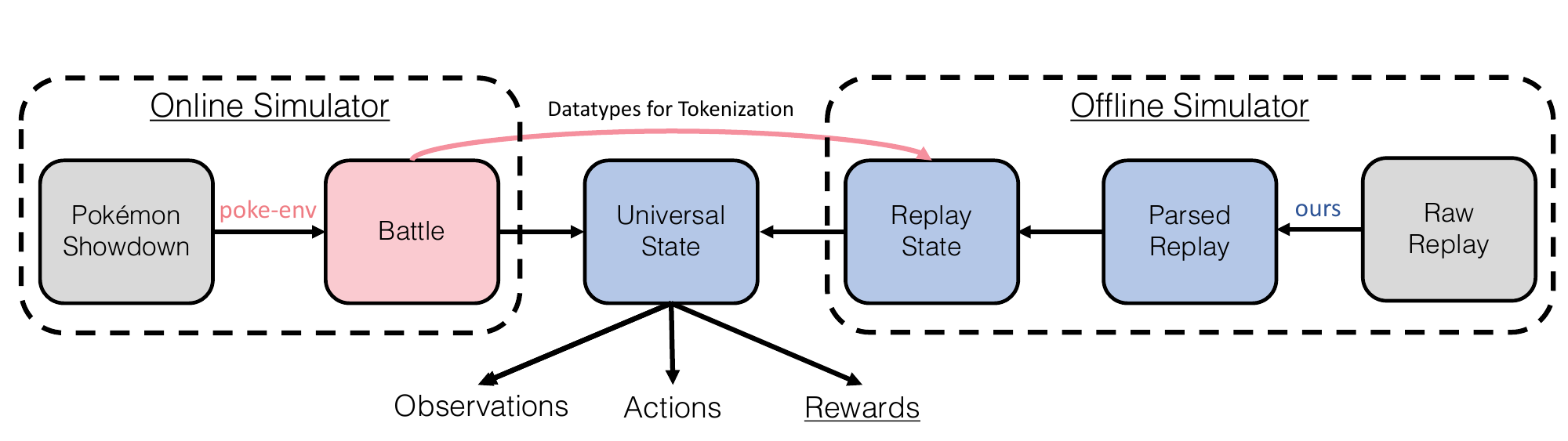}
    \caption{\textbf{Creating an Offline \texttt{poke-env} Dataset.} Our offline replay reconstruction pipeline interprets PS replays in a custom implementation designed to parse historical replays, improve team inference beyond the PS viewer, and diagnose failures. The resulting trajectory is then converted to a representation that can also be recovered from the online \texttt{poke-env} interface.}
    \label{fig:metamon_to_pokeenv_gap}
\end{figure}

\citet{ShowdownAICompetition} propose CPS and \poke Showdown as an important benchmark for AI research. \texttt{poke-env} \citep{poke-env} has made the PS domain much more accessible and has become the default for recent work, including ours and those in Appendix \ref{app:pokemon_llms}. Our \texttt{Metamon} release aims to be the final bridge connecting academic RL research and PS; we use a custom version of \texttt{poke-env} geared towards the early generations and add 1) a suite of additional baseline opponents, 2) standardized team sets, 3) a template for BC experiments, and 4) direct compatibility with large-scale RL training \citep{grigsby2024amago}. Fifth, and most importantly, we create the PS replay dataset with a complex reconstruction process (Section \ref{sec:dataset}). From the perspective of the user, our dataset appears to provide offline trajectories of human gameplay recorded via \texttt{poke-env}. At test time, the online \texttt{poke-env} interface is used to play against other agents and humans on the public ladder. However, this compatibility is an illusion enabled by closing a sim2sim gap between our own replay parser and \texttt{poke-env} (Figure \ref{fig:metamon_to_pokeenv_gap}). More discussion in Appendix \ref{app:replay_reconstruction}.

\subsection{Large Langauge Model Agents}

\label{app:pokemon_llms}
 
\poke\hspace{-1mm}'s web presence lets large language models (LLMs) act on \poke game states that involve many categorical variables that can be formatted as natural language. PokéLLMon \citep{hu2024pokellmon} conditions the LLM on a history of observations, actions, and turn results to select the next action. They also use retrieval-augmented generation from a \poke knowledge database to inform the LLM's decisions. PokéLLMon achieves a 49\% win rate on the Gen8RandomBattles \poke Showdown ladder but does not report Glicko-1 or GXE statistics that control for matchmaking bias. Note that the expected raw win rate on the \poke Showdown ladder in modern generations (where large player pools allow for even matchmaking) is $\approx\hspace{-1mm}50\%$ unless well below human-level. \citet{karten2025pokechamp} extend the LLM prompting setup to model the opponent's decisions and enable depth-limited search with heuristic value functions. Prompts include information like move damage calculations that let future outcomes inform action selection. Pokéchamp's planning allows for a $76\%$ win rate against PokéLLMon in Gen8RandomBattles and Gen9OU. Pokéchamp is concurrent work, and non-trivial modifications are needed to convert its model-based search to the gameplay mechanics of the early generations. Finally, \citet{ASimpleFrameworkforIntrinsicReward-ShapingforRLusingLLMFeedback} use an LLM for reward design to improve sample efficiency of DQN \citep{mnih2015human} against heuristics.

\section{Broader Related Work}

\label{app:general_related_work}

\textbf{Imitation Learning and Offline RL.} Many large-scale agents in complex domains are trained by imitation learning. These methods prioritize training scalable sequence models on large datasets and avoid RL obstacles \citep{reed2022generalist, gallouedec2024jack, jiang2022vima, raad2024scaling, brohan2023rt}. Offline RL \citep{prudencio2023survey, levine2020offline} learns policies that outperform their demonstrations and has found success at scale \citep{kumar2022offline, springenberg2024offline}. In practice, offline RL can be used in a way-off-policy or multi-batch setting \citep{laroche2019multi, najibiterative} where models are iteratively retrained or fine-tuned as data accumulates or better training techniques are found \citep{lampe2024mastering, tirumala2023replay}; the ability to learn stable policies from large mixed-quality datasets unlocks a flexible engineering workflow. Many solutions prevent the learned policy from deviating too far from the offline dataset \citep{wang2020critic, nair2020awac, fujimoto2021minimalist}. These approaches create a spectrum between unconstrained RL and behavior cloning and let a single objective replace the two-stage process of BC pre-training $\rightarrow$ RL fine-tuning.

Offline RL targets real-world use cases where (1) data collection is expensive or (2) deployment mandates some minimum performance standard well above random exploration. Playing \poke against humans leads to both problems: battles are slow (and there are limits to how many games we can play in parallel), and finding competent strategies across the full range of \poke teams and game modes is a daunting exploration challenge. In simulated RL domains, it is common to mimic the process of learning from existing data by first training online RL agents and then saving their rollouts for offline research \citep{reed2022generalist, fu2020d4rl, gulcehre2020rl, agarwal2020optimistic}. If online RL cannot solve the task, it may be possible to crowdsource demonstration datasets \citep{o2024open, gerstgrasser2022crowdplay}. It would be more realistic (and more convenient) if offline datasets already existed and grew naturally without requiring researchers to collect data. Our \poke dataset falls in this category --- as do other games played on the internet like Chess \citep{lichess2025}, Go \citep{kgsgo2025, silver2016mastering}, Diplomacy \citep{meta2022human}, and Starcraft II \citep{vinyals2019grandmaster, mathieu2023alphastar}. Other examples include autonomous driving \citep{kiran2021deep, lee2024ad4rl} and e-commerce \citep{saito2020open}.

\textbf{Gameplaying.} Games have always been key benchmarks for AI and RL research \citep{CAMPBELL200257, 10.1145/203330.203343}. High-profile successes include AlphaZero in chess and Go \citep{silver2018general}, AlphaStar in StarCraft II \citep{vinyals2019grandmaster}, OpenAI Five in DOTA 2 \citep{berner2019dota}, and DeepNash in Stratego \citep{perolat2022mastering}. Applications of model-based search to imperfect information games (IIGs) like poker \citep{moravvcik2017deepstack, brown2018superhuman, brown2020combining} create methods at the intersection of RL and game theory. We refer interested readers to \citet{schmid2023student} for a detailed overview. Policy learning in CPS (Section \ref{sec:method}) could also be viewed from the perspective of IIG formalisms like Factored Observation Stochastic Games \citep{schmid2021search}. Model-free RL against hybrid populations of opponent agents is a viable alternative despite lacking theoretical guarantees to converge to optimal (equilibrium) policies \citep{vinyals2019grandmaster, rudolph2025reevaluating, heinrich2015fictitious}. Finally, long-context sequence models have been used to model the decisions of opponents \citep{nashed2022survey} in multi-agent settings \citep{jing2024towards}.

\section{Heuristic Opponents}
\label{app:heuristics}

\begin{figure}[h!]
    \centering
    \begin{minipage}{0.53\textwidth}
        \centering
        \includegraphics[width=\linewidth]{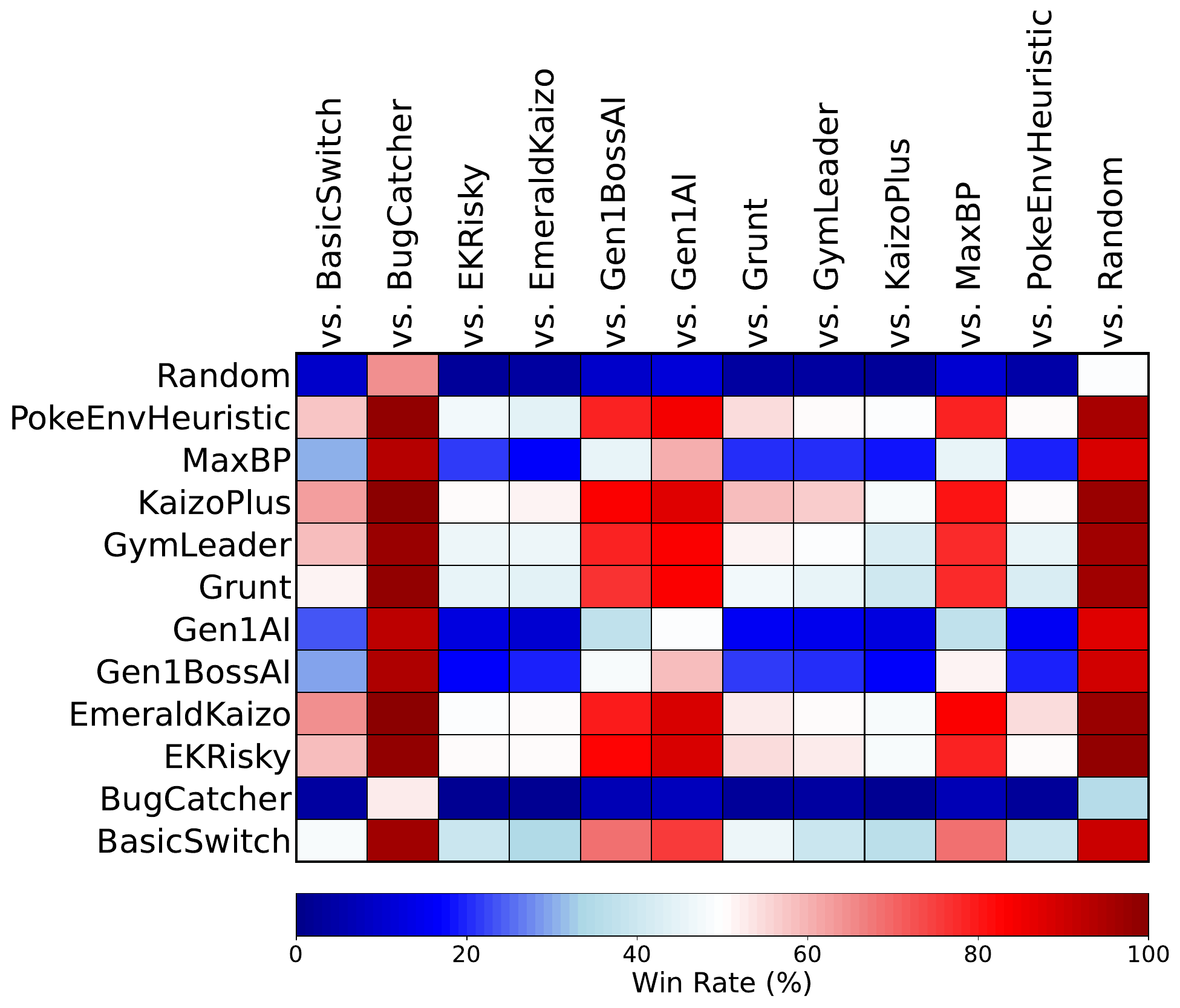}
        \caption{\textbf{Heuristic Round-Robin.} Entries denote the win rate of the row player against the column player in $2$k Variety Set battles across Gen1-4OU.}
        \label{fig:heuristic_round_robin}
    \end{minipage}
    \hfill
    \begin{minipage}{0.40\textwidth}
        \centering
        \includegraphics[width=\linewidth]{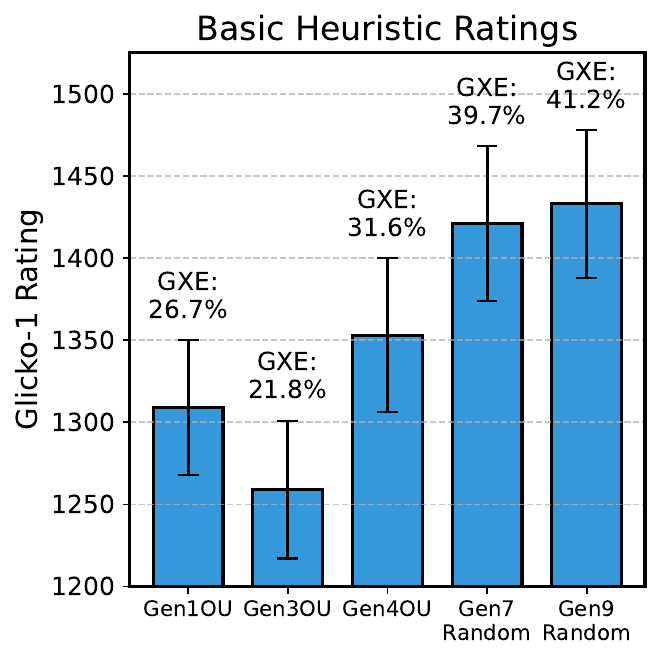}
        \caption{\textbf{PokeEnv Heuristic vs. Humans}. Early-Gen OU tiers are unique games that prioritize long-horizon control over memorization of damage matchups between \pokelower\hspace{-1mm}.}
        \label{fig:poke_env_heuristic_on_the_ladder}
    \end{minipage}
\end{figure}

In an attempt to evaluate a variety of \poke fundamentals, we develop an array of heuristic opponents. These policies are unable to cheat by accessing unrevealed information about their opponent's team but are otherwise free to use ground-truth knowledge of \poke\hspace{-1mm}'s mechanics to select actions. Figure \ref{fig:heuristic_round_robin} summarizes the relative performance of these heuristics. Ultimately, we find it difficult to generate meaningful diversity from this larger set and focus on six heuristics: 

\begin{itemize}
    \item \textbf{RandomBaseline} selects a legal move (or switch) uniformly at random and measures the most basic level of learning early in training runs. 
\end{itemize}
\begin{itemize}
    \item \textbf{Gen1BossAI} emulates the decision-making of opponents in the original \poke Generation 1 games. It usually chooses random moves. However, it prefers using stat-boosting moves on the second turn and ``super effective'' moves when available.
    \item \textbf{Grunt} is a maximally offensive player that selects the move that will deal the greatest damage against the current opposing \poke using \poke\hspace{-1mm}'s damage equation and a type chart and selects the best matchup by type when forced to switch. Its strategy amounts to greedy one-ply search and is an improvement over a common ``MaxBasePower'' agent in related work. 
    \item \textbf{GymLeader} improves upon Grunt by additionally taking into account factors such as health. It prioritizes using stat boosts when the current \poke is very healthy, and heal moves when unhealthy.
    \item \textbf{PokeEnv} is the \texttt{SimpleHeuristicsPlayer} baseline provided by \citet{poke-env}.
    \item \textbf{EmeraldKaizo} is an adaptation of the AI in a \poke Emerald ROM hack intended to be as difficult as possible. The game's online popularity has led to a community effort to document its decision-making in extensive detail. We use this documentation to re-implement the policy. It selects actions by scoring the available options against a rule set that includes handwritten conditional statements for a large portion of the moves in the game.
\end{itemize}

Figure \ref{fig:poke_env_heuristic_on_the_ladder} evaluates the PokeEnv Heuristic against humans on the ladder. We choose PokeEnv for this task because it appears in external work, but its strengths and weaknesses are similar to several other heuristics in our set. We use the same Competitive Team Set as our main model evaluations but evaluate over a smaller sample of battles per ruleset. The relationship between battle format and heuristic performance in Fig. \ref{fig:poke_env_heuristic_on_the_ladder} is predictable given knowledge of the PS metagames. Players correctly accuse the heuristic of being a bot in the online chat, and we decide we have made our point and stop evaluations. Notably, these accusations are not rooted in the super-human reaction time of the policy, but in its lack of move diversity and multi-turn strategy while playing at the (low) level people have come to expect from hobbyist bot projects and the \poke video games. Our learning-based agents do not suffer from these problems, and we will return to this discussion in Appendix \ref{app:experiment_details}. We would expect the heuristic to perform worst (and play least like a human) in Gen2OU, but are not comfortable evaluating this. Glicko-1 ratings can be slow to converge when this far below the mean, and it is possible that Fig. \ref{fig:poke_env_heuristic_on_the_ladder} is an overestimate. However, our low rating skews matchmaking in our favor (we are matched against the lowest ELO players) --- making this a rare case where raw win-loss records can be informative as an upper bound on win rate (Table \ref{tbl:poke_env_win_rates}).

\vspace{-3mm}

\begin{table}[h!]
\centering
\resizebox{.33\textwidth}{!}{%
\begin{tabular}{@{}lcc@{}}
\toprule
\textbf{Battle Format} & \textbf{Wins} & \textbf{Losses} \\ \midrule
Gen1OU                 & 16            & 59              \\
Gen3OU                 & 16            & 54              \\
Gen4OU                 & 21            & 36              \\
Gen7RandomBattle       & 24            & 32              \\
Gen9RandomBattle       & 28            & 32              \\ \bottomrule
\end{tabular}}
\caption{\textbf{PokeEnv Heuristic Win-Loss Records on the PS Ladder.}}
\label{tbl:poke_env_win_rates}
\end{table}

\section{Replay Reconstruction}
\label{app:replay_reconstruction}

As mentioned in Appendix \ref{app:rl_in_pokemon}, we build a custom replay reconstruction pipeline designed to interpret years-old records of human gameplay and identify teams from a spectator point-of-view (POV). The resulting trajectories train offline policies that can be deployed online via \texttt{poke-env} \citep{poke-env}. 

\begin{figure}[h!]
    \centering
    \vspace{-3mm}
    \includegraphics[width=\linewidth]{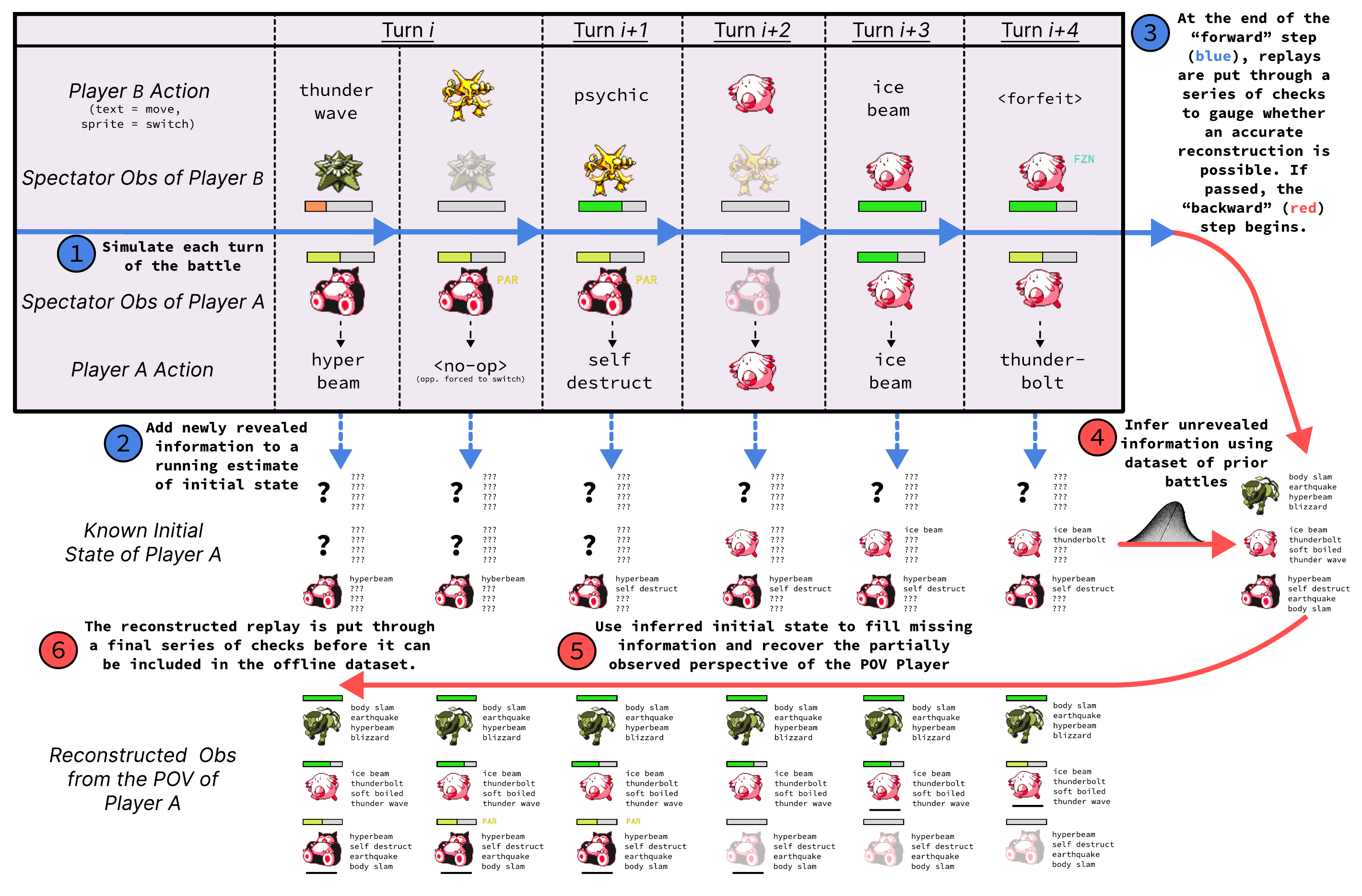}
    \vspace{-7mm}
    \caption{\textbf{Simplified Replay Reconstruction.} We walk through the reconstruction of the perspective of Player A in a Gen1OU example with teams of $3$ \poke\hspace{-1mm}.}
    \label{fig:main_text_replay_reconstruction}
\end{figure}

We follow a process visualized by a simplified example in Figure \ref{fig:main_text_replay_reconstruction} to extract complete battle information. On each turn, we add newly revealed information to a running estimate of the initial team configuration. By the end of the battle, some details may still be missing and are inferred with \poke Showdown statistics. We then backfill the inferred team through the trajectory, accounting for any changes to the roster that occur during the battle. Since Player A should have full knowledge of their own team, but limited knowledge of Player B's, we save a trajectory from Player A's perspective by using the inferred version of Player A's private state and the original spectator POV of Player B's state.

During reconstruction, we will obtain: (1) the complete team composition for each player and (2) per-turn observations from one player's POV. We can use a real battle as an example. You can view the relay \href{https://replay.pokemonshowdown.com/gen4nu-776588848}{at this link}. Figure \ref{fig:appendix_raw_replay} gives a sample of the raw PS log for this replay, while Figure \ref{fig:reconstructed_replay_team} shows a chosen POV player's observed and inferred team. Finally, Figure \ref{fig:appendix_parsed_replay} shows the fully reconstructed replay containing all necessary information for model training.

\subsection{Reconstruction Failures}
\label{app:replay_reconstruction:flaws}

A challenge in the replay reconstruction process is that inaccurate team inference can create inaccurate records of human decision-making: An expert player may have may have only picked the action in the replay because they did \textit{not} have access to the moves or \poke our dataset says they did. This is a fundamental problem created by the spectator POV, but it could be improved by team inference strategies that are more sophisticated than sampling from historical statistics.
\begin{wrapfigure}{r}{.42\textwidth}
\begin{center}
    \vspace{-5mm}
    \includegraphics[width=\linewidth]{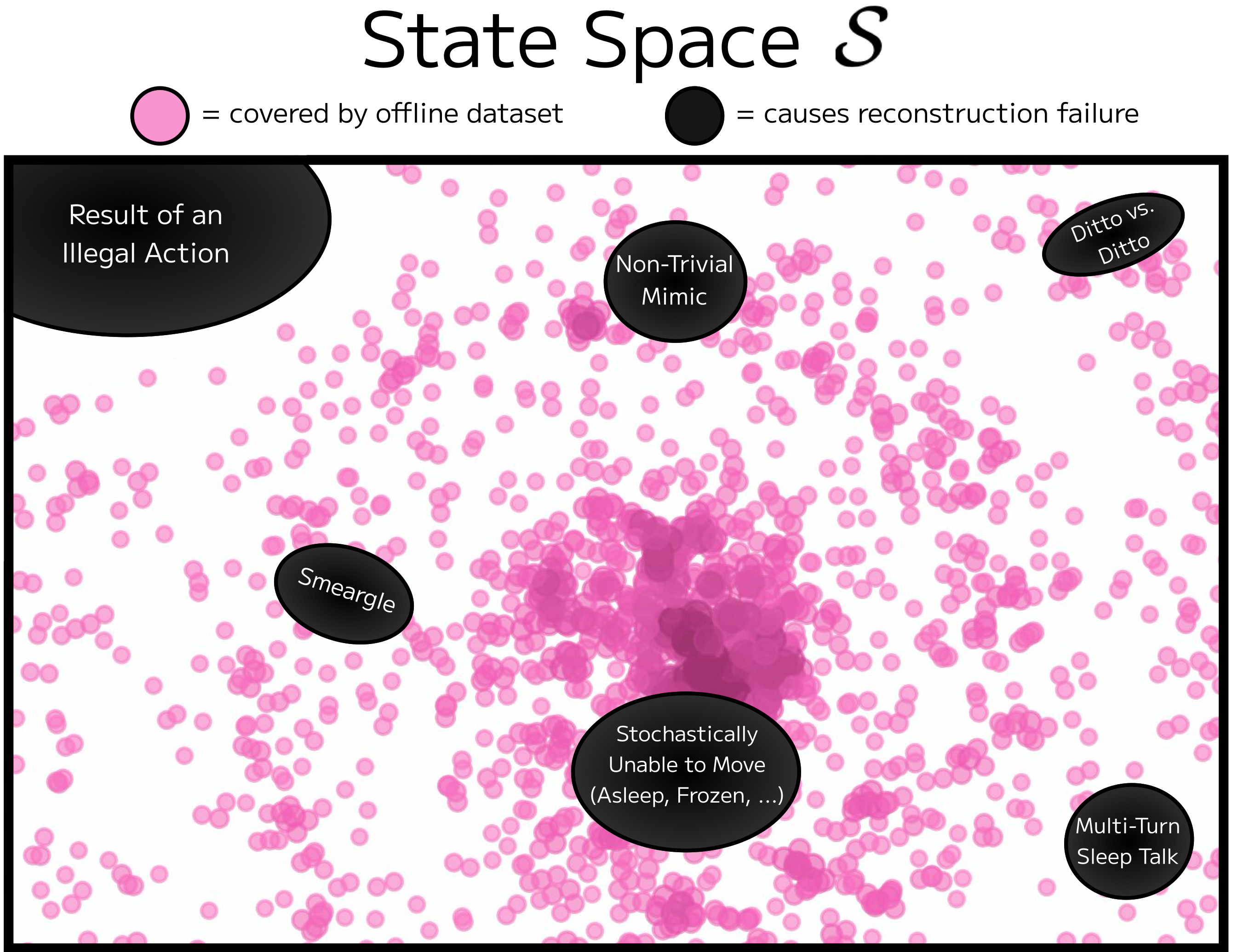}
    \vspace{-5mm}
    \caption{\textbf{Replay Parser Failure States.} An informal visualization of how the replay reconstruction process creates holes in our dataset on top of the more standard distribution shift inherent to offline RL.}
    \label{fig:state_space_holes}
    \vspace{-8mm}
\end{center}
\end{wrapfigure}

Offline RL always confronts a distribution shift problem created by sampling a finite dataset from a large state/action space \citep{levine2020offline}. Replay reconstruction can fail, and these failures add an additional challenge in that some specific state/actions will never appear in a dataset of any size (Figure \ref{fig:state_space_holes}). Some of these failures are caused by unimplemented game mechanics that rarely occur but could be improved. Others are caused by fundamentally ambiguous situations from the spectator perspective --- even the PS browser replay viewer gets these wrong or warns that values may be inaccurate. A long list of checks throughout the reconstruction process attempts to find and discard trajectories in these states. These situations are rare and discarding them may be needlessly cautious. 

There are two gaps in the replay dataset that we cannot ignore. Our solutions impact our findings and are worth discussing in detail:

\textbf{Illegal Actions.} \poke always has \textit{up to} $9$ discrete actions, but some of these actions become invalid as the battle progresses. Humans are not given the option to select invalid actions, so they never appear in the dataset. Offline RL should be able to handle this problem. Our policies are clearly told which actions are invalid, and we let their mistakes become indicators of accumulating OOD behavior\footnote{For reference, all RL policies average valid action rates of $97$-$99\%$ against heuristics and $95$-$98\%$ against humans. Nearly all of these invalid actions occur in succession once the policy is already in a lost position or runs into a limitation of the observation space discussed in Appendix \ref{app:space_design}.}. We send a random valid action to PS if an invalid action is selected. Invalid action masking was added to our open-source release long after the experiments in this paper and predictably made little difference when enforced only at test time --- though it may improve value estimation during training.

\begin{wrapfigure}{r}{.55\textwidth}
\begin{center}
    \vspace{-8mm}
    \includegraphics[width=\linewidth]{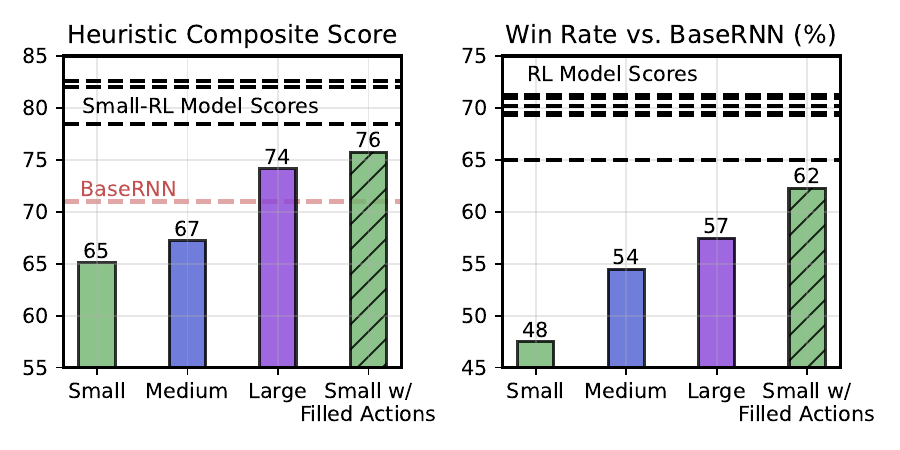}
    \vspace{-7mm}
    \caption{\textbf{Impact of Improved Missing Action Labels on a $15$M Transformer IL Policy.}}
\label{fig:filled_actions}
    \vspace{-6mm}
\end{center}
\end{wrapfigure}
\textbf{(Stochastically) Unrevealed Moves.} There are situations where the player's action choice has no impact on the battle and is not revealed to spectators. These occur too often to discard the trajectory, so we either need to mask or fill the action label. BC-RNN baselines (Appendix \ref{app:experimental_details:bcrnn}) \textit{mask} unrevealed action labels. When training offline RL along a full trajectory sequence in parallel, it is risky to back up the Q-values of timesteps where the actor or critic is not trained. Therefore, the RL models \textit{fill} action labels and the main Transformer BC models follow suit to create a direct comparison between variants of the actor objective (Eq. \eqref{eq:actor_objective}). We initially fill missing actions with a small BC-RNN model trained on a much earlier version of the dataset. The precise accuracy of these moves may not seem important because they have no impact on the battle. However, there are stochastic gameplay mechanics (mainly sleep and paralysis) where they \textit{could have} impacted the battle. We eventually suspect we can improve by filling missing actions with the more accurate (Figure \ref{fig:early_il_results}) BaseRNN model. We retrain $15$M IL and RL Transformer policies on this revised (``Filled Action'') version of the dataset. Offline RL should have already been able to avoid sub-optimal action choices in the situations they are relevant. Indeed, we find no evidence that the new action labels impact the RL policies. However, the Small IL model is significantly improved --- now ranking \textit{between} Large IL and the RL eval scores against heuristics (Figure \ref{fig:filled_actions} Left), and BC-RNN (Fig. \ref{fig:filled_actions} Right). Though not included in the figures, Small IL with Filled Actions also ranks between Large IL and all RL scores against Large IL (Figure \ref{fig:vs_large_il}) and the Foul Play engine (Figure \ref{fig:vs_foul_play}). 

We conclude that while the comparisons between IL and RL remain a fair evaluation of the same architecture trained on the same dataset, the original dataset was challenging in a way that was unintentionally similar to contrived benchmarks that dilute high-quality demonstrations with poor decisions \citep{fu2020d4rl}. Our final batch of RL models (SynRL-V1+SP, SynRL-V1++, and SynRL-V2) use improved labels in their human battle trajectories out of caution. After the release of this paper, we added missing action masking directly into the RL training pipeline with similar results.

\newpage

\begin{figure}[h!]
    \centering
    \includegraphics[width=\linewidth]{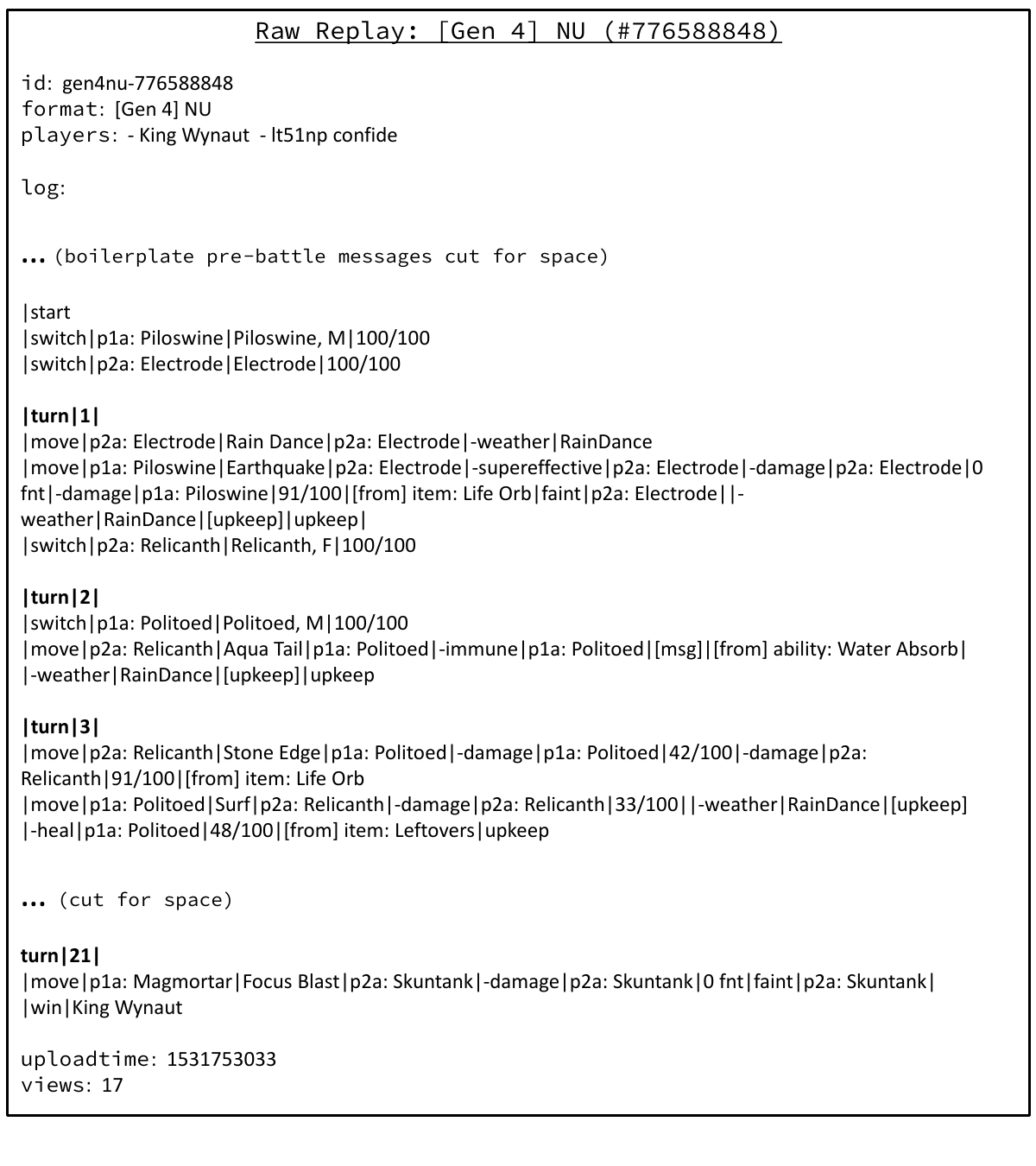}
    \caption{An example Gen4 NeverUsed (NU) replay file downloaded from PS server.}
    \label{fig:appendix_raw_replay}
\end{figure}

\newpage

\begin{figure}[h!]
    \centering
    \includegraphics[width=0.65\linewidth]{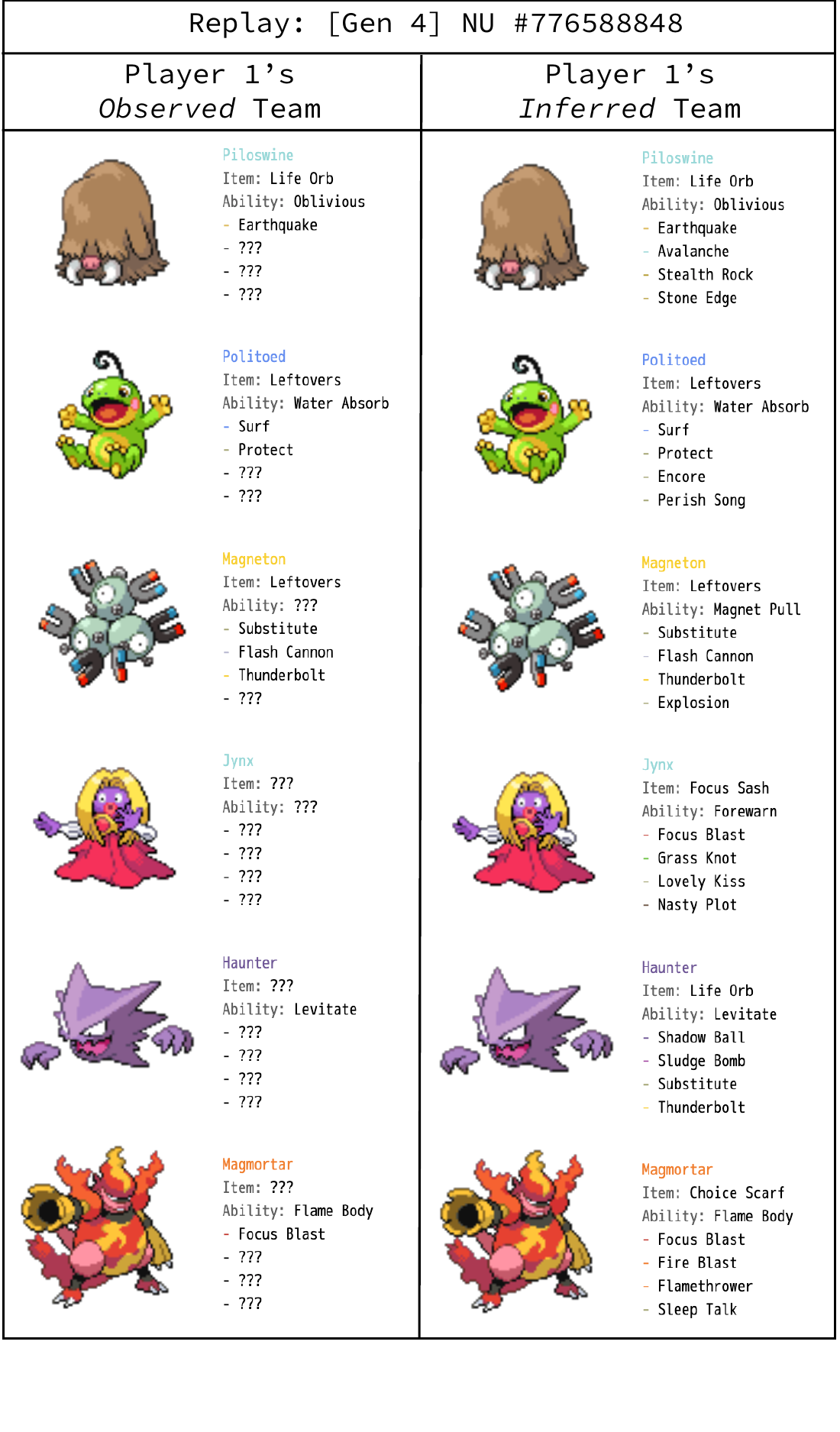}
    \caption{Continuing the Gen4 NU example by listing the observed team and the inferred team after replay reconstruction.}
    \label{fig:reconstructed_replay_team}
\end{figure}

\newpage

\begin{figure}[h!]
    \centering
    \includegraphics[width=\linewidth]{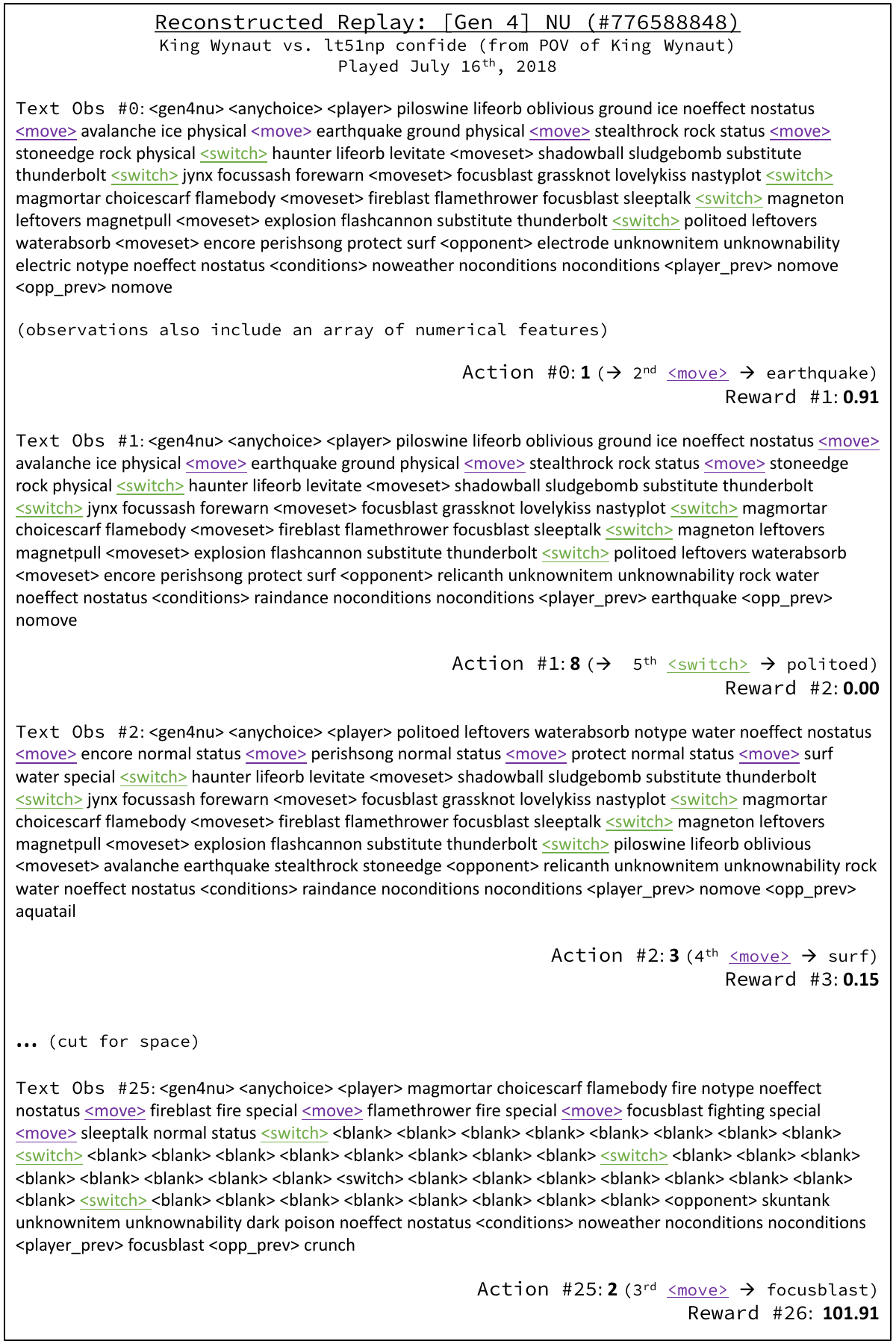}
    \caption{Concluding the Gen4 NU example with an abridged version of the reconstructed replay.}
    \label{fig:appendix_parsed_replay}
\end{figure}

\newpage

\section{Training Details}
\label{app:training_details}

\subsection{Reward Function}
\label{app:space_design}

Rewards are a combination of three shaping terms and a binary win/loss indicator ($r_{\text{win}}$):

\vspace{-8mm}
\begin{align*}
R(s_t, a_t)  = r_{\text{hp}} + \frac{1}{2} r_\text{{stat}} + r_{\text{faint}} + 100 r_{\text{win}}
\label{eq:reward}
\end{align*}
\vspace{-6mm}

We describe the shaping terms from the perspective of the agent's player:

\begin{itemize}
    \item  \textbf{Health Reward $r_{\text{hp}}$:} Encourages dealing more damage than the opponent and/or recovering more health than the opponent. Computed by net health points gained/lost by our active \poke versus those gained/lost by the opponent's active \poke\hspace{-1mm} (with all health values scaled $0-1$).
    \item \textbf{Status Reward $r_{\text{stat}}$:} Encourages dealing status conditions while avoiding taking status conditions ourselves. Status conditions are a key indicator of mid-game progress. Computed by the net gain in the binary presence of a status condition of the two \poke on the field. 
    \item \textbf{Faint Reward $r_{\text{faint}}$:} Encourages knocking out the opponent's \poke while preserving our own. Computed by the number of \poke we made unavailable to the player on this turn minus the number we lost.
\end{itemize}

The reward function is designed to give some shaping to help the offline filter $w$ (Equation \eqref{eq:actor_objective}) learn to assign unique weights over short horizons but be dominated by the binary win/loss outcome we ultimately care about. We do find some qualitative evidence of models exploiting the shaped terms. For example, our agents tend to cling to life in clearly lost positions by using recovery moves.

\subsection{Observation Space}
Observations include a language description (depicted by Figure \ref{fig:annotated_obs}) and $48$ numerical features. Numerical features include the base power and accuracy of moves and the health/stats/boosts of \poke\hspace{-1mm}. We defer a full account to the open-source release. Implementation details add the previous action and reward as policy inputs. Rewards may help resolve some ambiguity over the outcome of the previous turn (e.g., did the move hit and deal damage?). The player's previous action is a one-hot vector that is mostly redundant to information in the text observation but helps provide a history of action choices that were not revealed to the opponent.

Our observation space relies on long-term memory to track the true state of the battle. Section \ref{sec:method} notes that we only include the visible attributes of the opponent's active \poke\hspace{-1mm}, which reduces dimensionality and distribution shift over the opponent's team. We can infer the public state of the opponent's team from memory over the active \poke on previous turns and their move choices. Text tokens include the most recent move of both \poke on the field. The long-term memory of our models is quite effective in general. As one example, PS enforces a rule called ``Sleep Clause'' where attempting to put a second opponent \poke to sleep does nothing and wastes a turn. Our policies are remarkably good at following this rule even though their only way to track it is to recall that they put a \poke to sleep and that it has not reappeared and woken up.

There is a limit to the number of times a move can be used by a \poke in a battle. These ``PowerPoint'' (PP) limits break long stalemates in CPS, but PP counts are unreliable and full of edge cases in replays. While PP counts are tracked during reconstruction to help discard replays, we ultimately exclude them from the observation space. We decided to protect against sim2sim gaps because we assumed our agents would have to be unrealistically skilled to survive long enough for PP limits to be relevant. Our final policies are actually strong enough that PP stall losses are their most noticeable flaw and the leading cause of invalid action selections (Appendix \ref{app:replay_reconstruction:flaws}). PP counts can be inferred from memory over the move history. However, this is challenging in practice, especially when lacking opponent policies that force PP stalls during self-play. SynRL-V2 does demonstrate some ability to play around PP limits.

The observation space can be improved to address specific gameplay mechanics and will be version controlled for future comparisons. However, environments as complex as CPS will always have nuanced partial observability and benefit from the flexibility of sequence model policies.

\subsection{Action Space}
Agents play with $9$ discrete actions. The first four indices correspond to the active \poke\hspace{-1mm}'s moves, and the remaining indices switch to the other \poke on the player's team. The correspondence between action index and move/switch choices is indicated by both the text and numerical observation --- which arrange their features in a consistent alphabetical order. As discussed in Appendix \ref{app:replay_reconstruction:flaws}, actions become invalid over the course of a battle. Invalid actions are also noted in the observation. If the agent selects an invalid action, it is replaced by a random valid action within the environment's transition dynamics.

\subsection{Models and Hyperparameters}

Table \ref{app:hparams} details the default training configuration for Small ($15$M), Medium ($50$M), and Large ($200$M) model sizes. Table \ref{tbl:models} lists changes for all the models and ablations mentioned in the paper and released in our open-source code.

% \usepackage{graphicx}
\begin{table}[h!]
\centering
\begin{tabular}{lccc}
\hline
                                      & \textbf{Small}      & \textbf{Medium}     & \textbf{Large}   \\ \hline
Learning Rate                         & \multicolumn{3}{c}{1e-4}          \\
Linear LR Warmup Steps                       & \multicolumn{3}{c}{1000}          \\
Target Critic $\tau$     & \multicolumn{3}{c}{0.004}         \\
TD Loss Coeff                         & \multicolumn{3}{c}{10}            \\
Grad Clip                             & \multicolumn{3}{c}{1.5}           \\
L2 Coeff                              & \multicolumn{3}{c}{1e-4}          \\
Batch Size                            & 32         & 40         & 48      \\ \hline
Actor Activation             & \multicolumn{3}{c} {Leaky ReLU} \\
Actor Layers                          & \multicolumn{3}{c}{2}             \\
Actor Hidden Dimension                & 300       & 400        & 512      \\ \hline
Agent Popart \citep{hessel2019multi}                          & \multicolumn{3}{c}{True}          \\
Critic Ensemble Size \citep{chen2021randomized}                    & \multicolumn{3}{c}{4}             \\
Critic Layers                    & \multicolumn{3}{c}{2}             \\
Critic Activation                   & \multicolumn{3}{c} {Leaky ReLU} \\
Critic Hidden Dimension                    & 300       & 400        & 512      \\ \hline
Turn Encoder Token Dim          & 100       & 100        & 160      \\
Turn Encoder Layers         & 3         & 3          & 5        \\
Turn Encoder Summary Tokens   & 4         & 6          & 11       \\
Turn Encoder Attention Heads     & 5         & 5          & 8        \\
Turn Encoder Numerical Tokens & \multicolumn{3}{c}{6}             \\ \hline
Causal Transformer Layers          & 3         & 6          & 9        \\
Causal Transformer Attention Heads           & 8         & 8          & 20       \\
Causal Transformer FF Dim.              & 2048      & 3072       & 5120     \\
Causal Transformer Model Dim.           & 512       & 768        & 1280     \\
NormFormer \citep{shleifer2021normformer}  & \multicolumn{3}{c}{True}          \\
$\sigma$Reparam \citep{zhai2023stabilizing}     & \multicolumn{3}{c}{True}          \\
Causal Transformer Normalization               & \multicolumn{3}{c}{LayerNorm \citep{ba2016layer}}       \\
Max Context Length & 200 & 200 & 128 \\
Causal Transformer Activation         & \multicolumn{3}{c}{Leaky ReLU} \\ \hline
\end{tabular}
\caption{\textbf{Base Training Hyperparameters by Model Size.} In reference to the architecture in Figure \ref{fig:architecture} and the AMAGO training configuration \citep{grigsby2024amago}.}
\label{app:hparams}
\end{table}

\begin{table}[h!]
\centering
\resizebox{\textwidth}{!}{%
\begin{tabular}{@{}llccl@{}}
\toprule
\multicolumn{1}{c}{\uline{\textbf{Model Name}}} &
\multicolumn{1}{c}{\uline{\textbf{Dataset}}} &
\multicolumn{1}{c}{\textbf{\begin{tabular}[c]{@{}c@{}}Architecture\\ (Table \ref{app:hparams})\end{tabular}}} &
\multicolumn{1}{c}{\textbf{\begin{tabular}[c]{@{}c@{}}Loss\\ (Equation \eqref{eq:actor_objective})\end{tabular}}} &
\multicolumn{1}{c}{\uline{\textbf{Notes}}} \\
\midrule

\textbf{Small IL} & RPS 950k & Small & Behavior Cloning & \\ \midrule

\textbf{Small IL (Filled Actions)} & RPS 950k & Small & Behavior Cloning & \begin{tabular}[c]{@{}l@{}}A late ablation that replaces \\ missing actions with improved \\ estimates (Appendix \ref{app:replay_reconstruction:flaws}).\end{tabular} \\ \midrule

\textbf{Small RL} & RPS 950k & Small & Exponential $w$ & \begin{tabular}[c]{@{}l@{}}Exp $w$ default to $\beta = .5$ \\ and clip weight values in $[1e$-$5$, 50].\end{tabular} \\ \midrule

\textbf{Small RL (Binary)} & RPS 950k & Small & Binary $w$ & \\ \midrule

\textbf{Small RL (Exp Extreme)} & RPS 950k & Small & Exponential $w$ & \begin{tabular}[c]{@{}l@{}}$\beta = 1$, clip $[-1e$-$5, 100]$. \\ (Testing sensitivity to $w$ hyperparameters).\end{tabular} \\ \midrule

\textbf{Small RL (Aug)} & RPS 950k & Small & Exponential $w$ & \begin{tabular}[c]{@{}l@{}}Randomly sets tokens to \\ \texttt{<unknown>} timestep-wise.\end{tabular} \\ \midrule

\textbf{Small RL (Filled Actions)} & RPS 950k & Small & Binary $w$ & \begin{tabular}[c]{@{}l@{}}Ablation that replaces missing actions \\ with improved estimates \\ (Appendix \ref{app:replay_reconstruction:flaws}).\end{tabular} \\ \midrule

\textbf{Small RL (Binary+MaxQ)} & RPS 950k & Small & Binary $w$, $\lambda = 1$ & \begin{tabular}[c]{@{}l@{}}Testing Q overestimation on the \\ replay dataset before scaling up.\end{tabular} \\ \midrule

\textbf{Medium IL} & RPS 950k & Medium & Behavior Cloning & \\ \midrule

\textbf{Medium RL} & RPS 950k & Medium & Exponential $w$ & \\ \midrule

\textbf{Medium RL (Aug)} & RPS 950k & Medium & Exponential $w$ & \\ \midrule

\textbf{Medium RL (Binary+MaxQ)} & RPS 950k & Medium & Binary $w$, $\lambda = 1$ & \\ \midrule

\textbf{Large IL} & RPS 950k & Large & Behavior Cloning & \begin{tabular}[c]{@{}l@{}}All Large architecture models use \\ the "Aug" dropout scheme by default.\end{tabular} \\ \midrule

\textbf{Large RL} & RPS 950k & Large & Exponential $w$ & \\ \midrule

\textbf{Large RL (Binary+MaxQ)} & RPS 950k & Large & Binary $w$, $\lambda = 1$ & \\ \midrule

\textbf{SyntheticRL-V0} & \begin{tabular}[c]{@{}l@{}}RPS 950k \\ + 1M Gen1\&3 \\ Variety Set model vs. model \\ trajectories with an ad-hoc \\ mixture of all policies above. \\ 2M total trajectories.\end{tabular} & Large & Binary $w$, $\lambda = 10$ & \\ \midrule

\textbf{SyntheticRL-V1} & \begin{tabular}[c]{@{}l@{}}SyntheticRL-V0 \\ + 1M Gen2 \& Gen4 \\ Variety Set model vs. model \\ battles. 3M total trajectories.\end{tabular} & Large & Binary $w$, $\lambda = 10$ & \\ \midrule

\textbf{SyntheticRL-V1+SelfPlay} & \begin{tabular}[c]{@{}l@{}}SyntheticRL-V1 \\ + 2M Gen1-4 SynRL-V1 \\ self-play trajectories. Models \\ from here to the end of the \\ table use improved action \\ labels in their RPS dataset \\ (Appendix \ref{app:replay_reconstruction:flaws}). \\ 5M total trajectories.\end{tabular} & Large & Binary $w$, $\lambda = 10$ & \\ \midrule

\textbf{SyntheticRL-V1++} & \begin{tabular}[c]{@{}l@{}}SyntheticRL-V1 \\ + 2M additional \\ Variety Set model vs. model \\ trajectories (5M total).\end{tabular} & Large & Binary $w$, $\lambda = 10$ & \\ \midrule

\textbf{SyntheticRL-V2} & \begin{tabular}[c]{@{}l@{}}SyntheticRL-V1++ adding \\ 100k trajectories from 50k \\ human battles in Jan-Mar \\ $2025$ (RPS 1.05M) --- \\ including many of our \\ own public battles. \\ 5M total trajectories.\end{tabular} & Large & Binary $w$ & \begin{tabular}[c]{@{}l@{}}Value predictions are converted to \\ two-hot classification following \\ \citep{grigsby2024amago2} \\ with $96$ output bins spaced \\ \textit{evenly} between $[-110, 110]$.\end{tabular} \\ 
\bottomrule
\end{tabular}%
}
\caption{\textbf{Model Variations}. Datasets, architectures, and hyperparameter changes (from the base set in Table \ref{app:hparams}) for the 20 main Transformer models trained throughout the paper. ``RPS 950k'' refers to the original replay reconstruction dataset (Appendix \ref{app:replay_reconstruction}). ``Exponential'' weight functions ($w$) are implemented following AWAC \citep{nair2020awac}. ``Binary'' weight functions are implemented following CRR \citep{wang2020critic}. In both cases, advantage estimates approximate $V(s)$ as the mean over the critic ensemble. ``Synthetic'' models increase batch size from $48 \rightarrow 96$ sequences.}
\label{tbl:models}
\end{table}

We train all models on a single $8 \times$ NVIDIA A$5000$ GPU machine for at least $1$M gradient steps. We default to the checkpoint at $1$M, which is well after performance has converged according to our evaluations. In the open-source code and weights, an ``epoch'' is an arbitrary interval of $25$k gradient steps, and we save checkpoints every $2$ epochs. Therefore, results default to checkpoint $40$ unless otherwise noted. SyntheticRL-V1+SelfPlay fine-tunes from epoch $40\rightarrow48$ and defaults to 48, while SyntheticRL-V2 is an exception in that we can confirm it is still improving at $1$M, and so we use the last available checkpoint (of $48$). These exceptions are noted by Table \ref{tab:model_usernames}, and Appendix \ref{app:experimental_details:self_play} contains more discussion.

\begin{figure}[h!]
    \centering
    \begin{minipage}{0.49\textwidth}
        \centering
        \includegraphics[width=\linewidth]{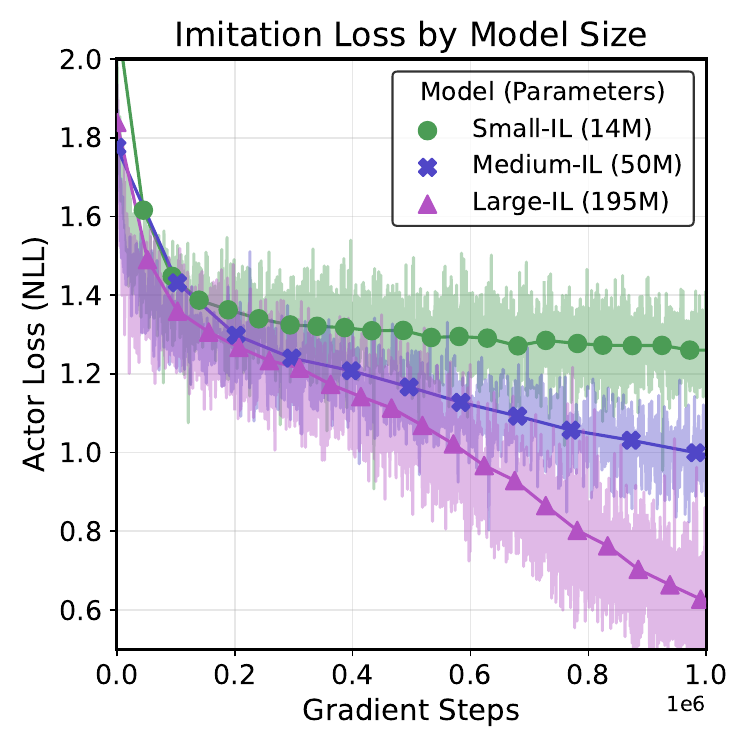}
        \vspace{-6mm}
        \caption{\textbf{Transformer IL Train Loss Curves.} Training loss on the \poke human replay dataset has a predictable relationship with model size when using a standard BC objective.}
        \label{fig:il_losses}
    \end{minipage}
    \hfill
    \begin{minipage}{0.49\textwidth}
        \centering
        \vspace{-3mm}
        \includegraphics[width=\linewidth]{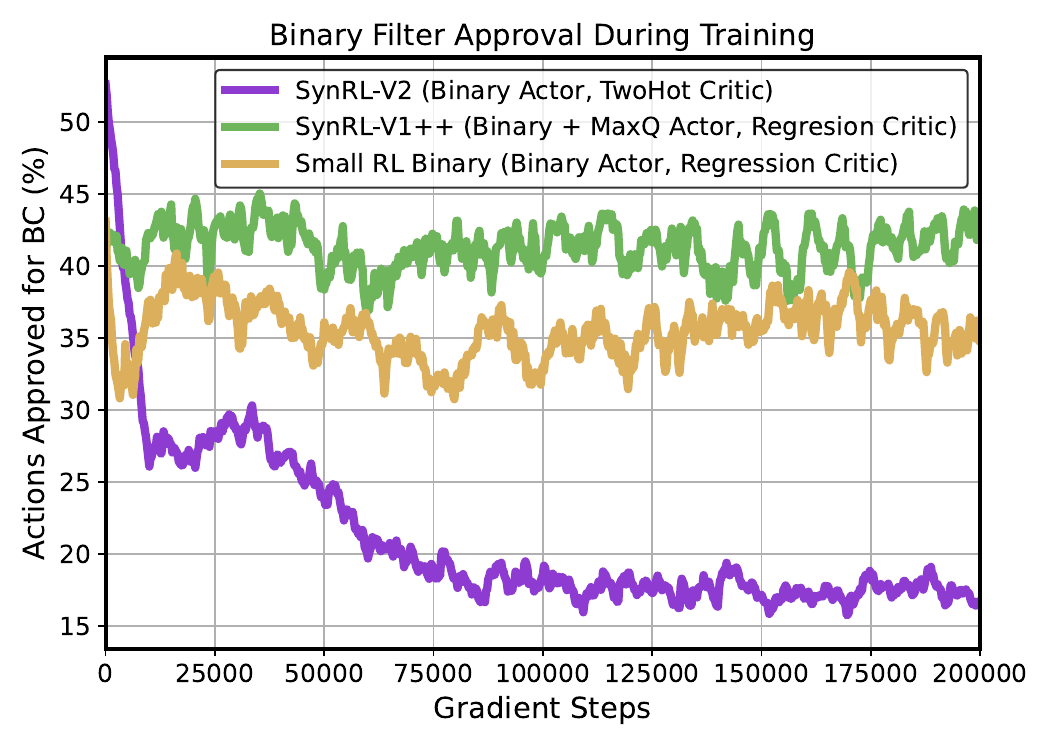}
        \caption{\textbf{Critic Filter Pessimism.} We track the percentage of the offline dataset assigned as weight $w(h, a) > 0$ (Eq. \eqref{eq:actor_objective}) throughout training. The accuracy of the two-hot classification filter has a significant impact on the pessimism of the BC process. Curves are noisy because they track the average value of a single GPU minibatch (of $12$ battles).}
        \label{fig:binary_filters}
    \end{minipage}
\end{figure}

Figure \ref{fig:il_losses} shows the relationship between model size and action prediction accuracy for behavior cloning models on the replay dataset. Figure \ref{fig:binary_filters} highlights the difference between scalar regression and two-hot classification for value prediction. 

\section{Experimental Details and Additional Figures}
\label{app:experiment_details}

This section contains figures and experimental details that support Section \ref{sec:experiments} in the main text.

\subsection{Early Imitation Learning Models}
\label{app:experimental_details:bcrnn}
\label{app:base_rnn}

\begin{figure}[h!]
    \centering
    \begin{minipage}{0.53\textwidth}
        \centering
        \vspace{-3mm}
        \includegraphics[width=\linewidth]{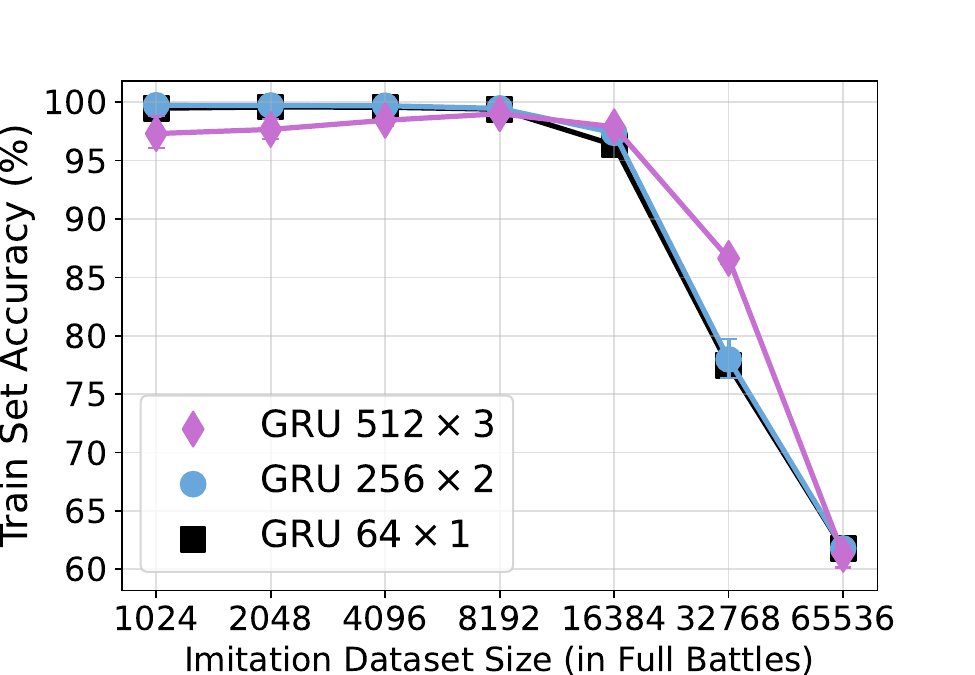}
        \caption{\textbf{Underfitting on PS Replays.} We report the train-set accuracy of (small) recurrent BC policies on increasingly large datasets of human gameplay. Error bars denote the maximum and minimum over four random subsets. Model sizes are reported by their hidden state and number of recurrent layers.}
        \label{fig:bc_rnn_underfitting}
    \end{minipage}
    \hfill
    \begin{minipage}{0.45\textwidth}
        \centering
        \includegraphics[width=\linewidth]{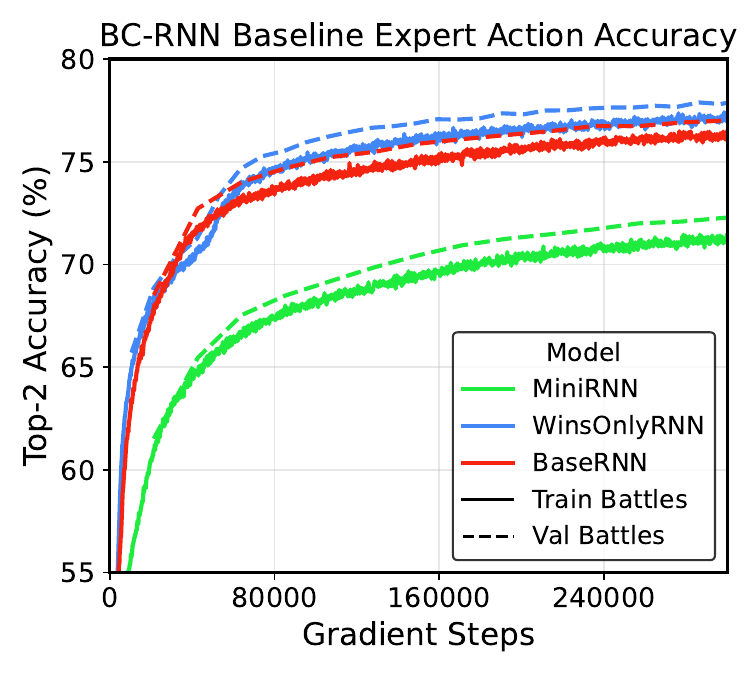}
        \vspace{-6mm}
        \caption{\textbf{BC-RNN Accuracy}. Action labels are high-entropy and we find Top-2 accuracy to be a more useful metric for tuning. ``BaseRNN'' is 3.5M params, ``MiniRNN'' ablates to $800$k, and ``WinsOnlyRNN'' follows the filtered BC approach of only imitating decisions from the POV of the winning player (cutting its train/val sets in half).}
        \label{fig:early_il_results}
    \end{minipage}
\end{figure}

In the beginning of our effort, it is not apparent that the \poke replay dataset requires architecture sizes beyond the scale of common RL problems. We begin by building a small-scale behavior cloning pipeline (that is still available in the \texttt{Metamon} code release). Figure \ref{fig:bc_rnn_underfitting} identifies clear underfitting on the reconstructed battle replay dataset. Our early development leads to the Turn Encoder Transformer architecture (Figure \ref{fig:architecture}) with a GRU-based \citep{cho2014learning} trajectory model (rather than the Transformer in Fig. \ref{fig:architecture}) to create a ``BaseRNN`` opponent. BaseRNN leads the early Heuristic Composite Score rankings (Figure \ref{fig:heuristic_composites}) and later serves as a fast (CPU-only) opponent and as a way to fill missing action labels (Appendix \ref{app:replay_reconstruction:flaws}). Figure \ref{fig:early_il_results} documents BaseRNN's predictive accuracy alongside two ablations. ``WinsOnlyRNN'' follows a common offline RL ablation by testing whether performance can be improved by manually discarding low-return trajectories from the POV of the losing player.

\subsection{Heuristic Evaluations}
\label{app:experimental_details:heuristics}

\begin{figure}[h!]
    \centering
    \includegraphics[width=\linewidth]{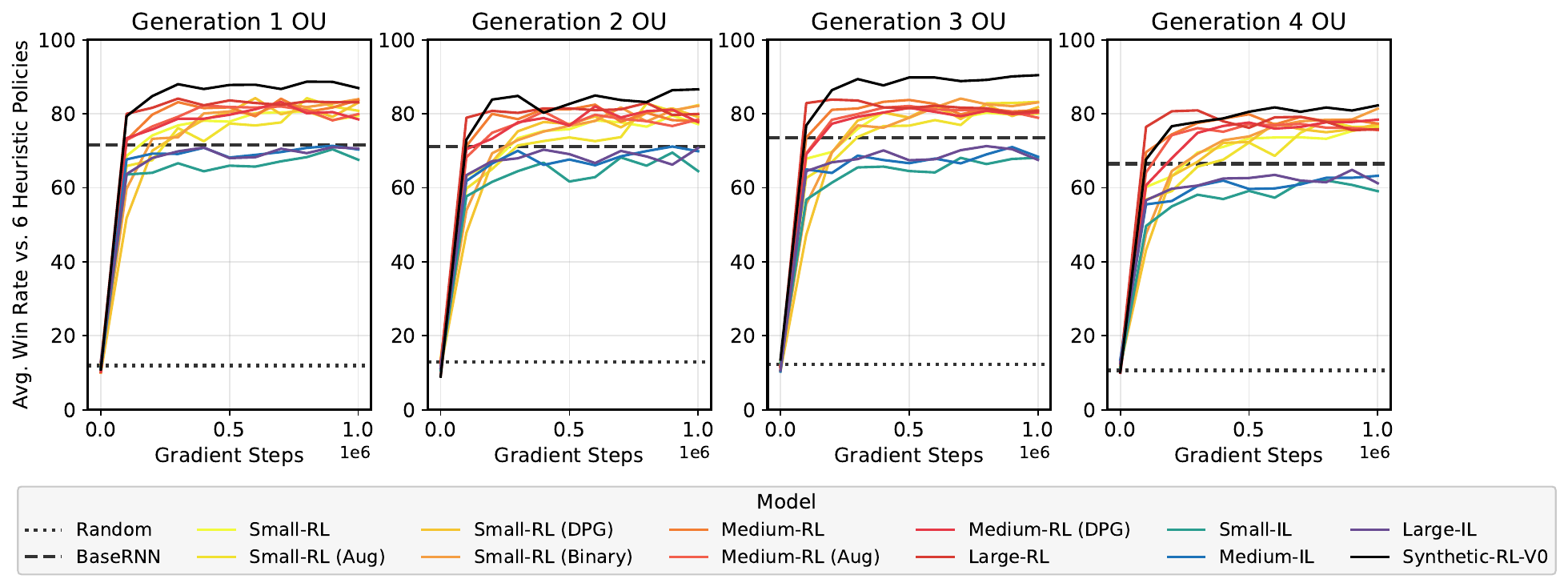}
    \caption{\textbf{Heuristic Composite Learning Curves.} Performance converges quickly but shows no sign of degrading over long training runs. BC and offline RL form two clear clusters with $\mathcal{L}_{\text{actor}}$ changes and model size having no clear impact.}
    \label{fig:heuristic_learning_curves}
\end{figure}

Figure \ref{fig:heuristic_learning_curves} records the Heuristic Composite Score (Section \ref{sec:experiments:heuristics}) of various models (Table \ref{tbl:models}) throughout training. Much of our early effort goes into creating strong but inexpensive heuristics to monitor training progress, but performance converges in less than $250$k training steps. Model-based opponent evaluations run fast enough to generate learning curves after the fact and shed more light on the relationship between training budget and performance (Appendix \ref{app:experimental_details:model_based}).

\subsection{Model-Based Evaluations}
\label{app:experimental_details:model_based}
\label{app:experimental_details:self_play}

\begin{figure}[h!]
    \centering
    \begin{minipage}{0.53\textwidth}
        \centering
        \includegraphics[width=\linewidth]{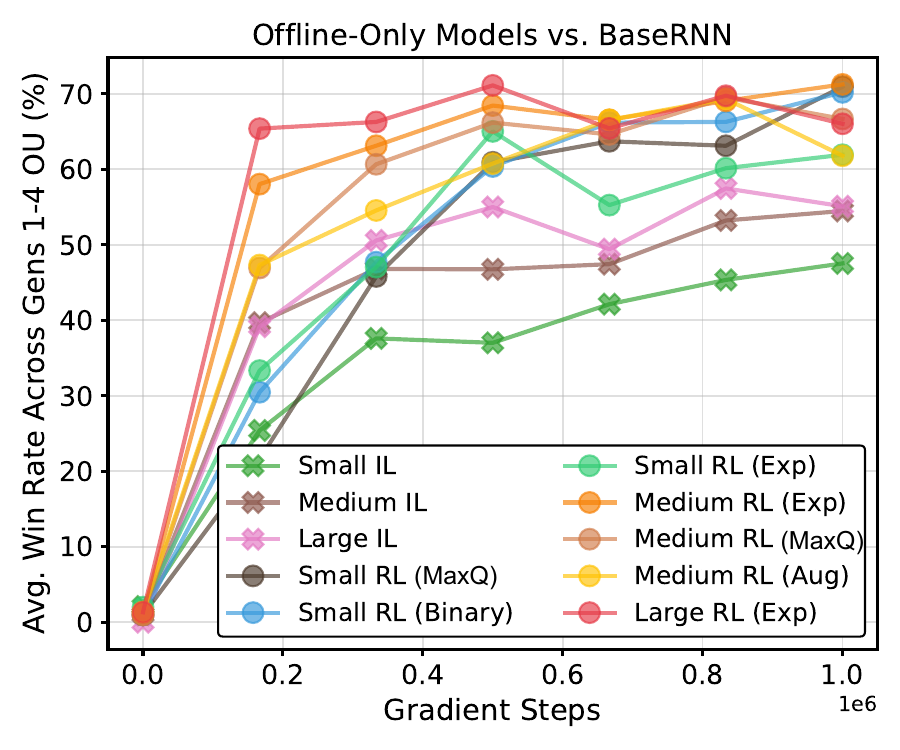}
    \caption{\textbf{Transformer IL and RL vs. RNN BC.} We evaluate the performance of Transformer policies trained on the offline replay dataset against a smaller RNN-based model designed for CPU-only inference. The RL updates do not display meaningfully distinct performance but outperform BC at all model sizes.}
    \label{fig:vs_basernn}
    \end{minipage}
    \hfill
    \begin{minipage}{0.45\textwidth}
        \centering
        \includegraphics[width=\linewidth]{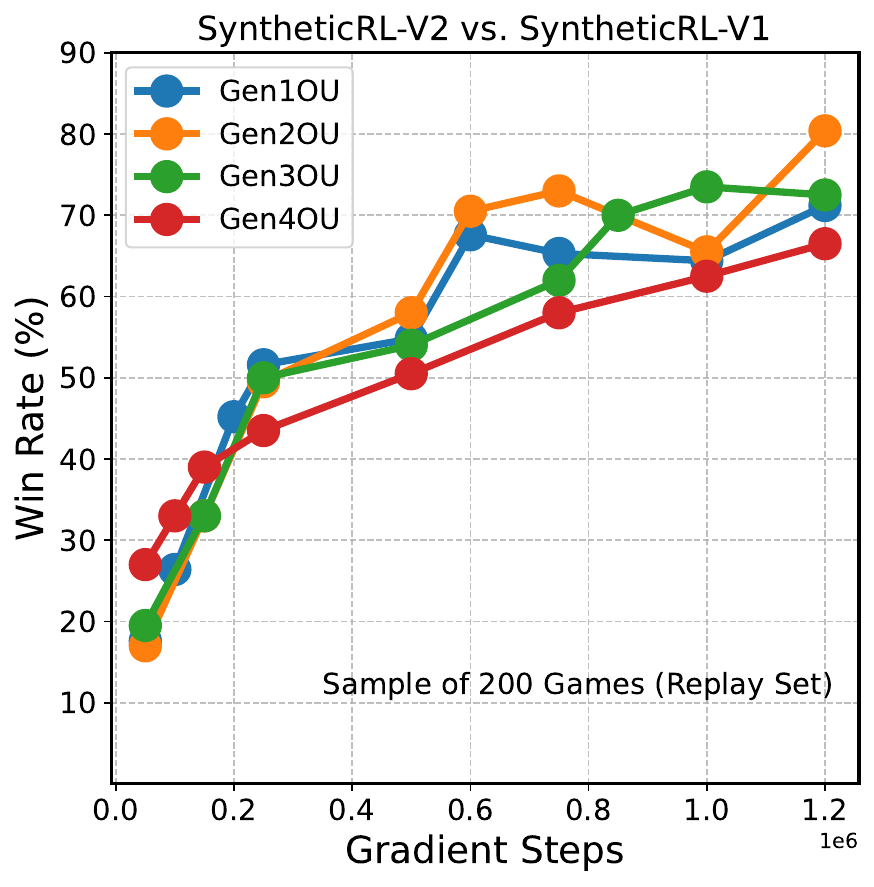}
    \caption{\textbf{Improvement of Advanced Policies.} We record the improvement of our best model (``SyntheticRL-V2'') against a previous version that reached a top $50$ ranking in Gen1OU.}
    \label{fig:synv2_vs_synv1}
    \end{minipage}
\end{figure}

Figure \ref{fig:vs_basernn} evaluates a variety of models against the BaseRNN behavior cloning model. Like the heuristic learning curve in Figure \ref{fig:heuristic_learning_curves}, performance converges well before the end of training against this opponent. Figure \ref{fig:synv2_vs_synv1} highlights the continued improvement of our final model (``SyntheticRL-V2``) against a previous version that had climbed into the global top $50$ in Gen1OU. SyntheticRL-V2 may not have converged after $1.2$M steps, but training was cut short due to time constraints. Table \ref{tab:syntheticRLV1V2} evaluates the impact of narrow self-play data with realistic teams and controls for the additional training budget of fine-tuning on this dataset versus continuing training on the original dataset.

\begin{table}[h!]
\centering
\resizebox{.6\textwidth}{!}{%
\begin{tabular}{ccccc}
\hline
                               & Gen1OU & Gen2OU & Gen3OU & Gen4OU \\ \hline
SyntheticRL-V1+SelfPlay @ 1.2M Steps & 63.6\%   & 59.6\%   & 61.4\%   & 59\%     \\
Synthetic-V1 @ 1.2M Steps             & 50\%     & 53.8\%   & 48.4\%   & 48.2\%   \\ \hline
\end{tabular}%
}
\caption{\textbf{Win Rates vs. SyntheticRL-V1.} We evaluate a checkpoint fine-tuned on a dataset of self-play battles against the original version (at $1$M training steps). We control for the additional training steps with a second version that maintains its original dataset. Sample size of $500$ games.}
\label{tab:syntheticRLV1V2}
\end{table}

\subsection{Human Evaluations}

Our models play under identical conditions to humans. We assign each model its own username (Table \ref{tab:model_usernames}). Usernames are visible to the opponent, so humans can adapt to the model over repeat matchups (just as they might exploit any other player). We use the PS statistics for each username in Figures \ref{fig:human_evals} and \ref{fig:human_percentiles}. Note that ratings like ELO and Glicko-1 confidence intervals decay every $24$ hours, so the PS statistics at the time of reading will no longer match our figures. Table \ref{tbl:raw_win_loss} records each model's overall win/loss for completeness --- though we note again that such records have little meaning because PS matches stronger models against stronger players. 

The PS ladder has increment time controls (similar to chess) that go into effect if requested by either player. We always request the timer in order to keep evaluations moving if our opponent disconnects from the game for an extended time. Note that most players also request time constraints, as Early Gen battles can be $20$+ minutes long even when enabled. Time limits can be a key constraint for CPS AI methods involving search or LLMs \citep{karten2025pokechamp}. However, our agents select an action at the inference speed of $\leq 200$M parameter Transformer, and this makes time constraints a non-issue. In fact, our pace of play is suspiciously fast. However, the opponent must be playing very quickly for this to be noticeable because decisions are made simultaneously, and the battle moves at the pace of the slower player. As we reach high ELO, we begin to run into the few players who can defeat our models while playing quickly. We eventually implement a random delay to hide the inference speed (while still playing faster than the opponent on most turns). Super-human speed aside, all our policies play in an undeniably human-like style. We saved hundreds of battle replays to the PS website, which you can browse via the links in Table \ref{tab:model_usernames} or by searching \url{https://replay.pokemonshowdown.com/}. These replays are a (mostly) unbiased sample of all matches played in public (spectator-viewable) battles while the lead author was monitoring the ladder evaluations. 

% \begin{table}[h!]
% \centering
% \resizebox{.4\textwidth}{!}{%
% \begin{tabular}{@{}cc@{}}
% \toprule
% \textbf{Model Name}                                                  & \textbf{PS Username} \\ \midrule
% Small-IL                                                             & \href{https://replay.pokemonshowdown.com/?user=SmallSparks}{SmallSparks}          \\ \midrule
% Large-IL                                                             & \href{https://replay.pokemonshowdown.com/?user=DittoIsAllYouNeed}{DittoIsAllYouNeed}    \\ \midrule
% Large-RL                                                             & \href{https://replay.pokemonshowdown.com/?user=Montezuma2600}{Montezuma2600}        \\ \midrule
% SyntheticRL-V0                                                       & \href{https://replay.pokemonshowdown.com/?user=Metamon1}{Metamon1}             \\ \midrule
% SyntheticRL-V1                                                       & \href{https://replay.pokemonshowdown.com/?user=TheDeadlyTriad}{TheDeadlyTriad}       \\ \midrule
% \begin{tabular}[c]{@{}c@{}}SyntheticRL-V1\\ + Self-Play\end{tabular} & \href{https://replay.pokemonshowdown.com/?user=ABitterLesson}{ABitterLesson}  \\ \midrule
% SyntheticRL-V1++ & \href{https://replay.pokemonshowdown.com/?user=QPrime}{QPrime} \\ \midrule
% SyntheticRL-V2 & \href{https://replay.pokemonshowdown.com/?user=MetamonII}{MetamonII} \bottomrule
% \end{tabular}%
% }
% \caption{\textbf{Public Ladder Usernames.} Models are tied to unique usernames throughout evaluations. Links lead to a replay page for each model. Miscellaneous test battles are also played under the usernames ``NotableWalrus'' and ``PsyduckIsUbers'', which are not always the same model and do not appear in results, but may be present in replays or videos featured in our release materials.}
% \label{tab:model_usernames}
% \end{table}

\begin{table}[h!]
\centering
\resizebox{.5\textwidth}{!}{%
\begin{tabular}{@{}ccc@{}}
\toprule
\textbf{Model Name} & \textbf{PS Username} & \textbf{Checkpoint} \\ \midrule
Small-IL & \href{https://replay.pokemonshowdown.com/?user=SmallSparks}{SmallSparks} & 40 \\ \midrule
Large-IL & \href{https://replay.pokemonshowdown.com/?user=DittoIsAllYouNeed}{DittoIsAllYouNeed} & 40 \\ \midrule
Large-RL & \href{https://replay.pokemonshowdown.com/?user=Montezuma2600}{Montezuma2600} & 40 \\ \midrule
SyntheticRL-V0 & \href{https://replay.pokemonshowdown.com/?user=Metamon1}{Metamon1} & 40 \\ \midrule
SyntheticRL-V1 & \href{https://replay.pokemonshowdown.com/?user=TheDeadlyTriad}{TheDeadlyTriad} & 40 \\ \midrule
\begin{tabular}[c]{@{}c@{}}SyntheticRL-V1\\ + Self-Play\end{tabular} & \href{https://replay.pokemonshowdown.com/?user=ABitterLesson}{ABitterLesson} & 48 \\ \midrule
SyntheticRL-V1++ & \href{https://replay.pokemonshowdown.com/?user=QPrime}{QPrime} & 40 \\ \midrule
SyntheticRL-V2 & \href{https://replay.pokemonshowdown.com/?user=MetamonII}{MetamonII} & 48 \\ \bottomrule
\end{tabular}%
}
\caption{\textbf{Public Ladder Usernames.} Models are tied to unique usernames throughout evaluations. Links lead to a replay page for each model. Miscellaneous test battles are also played under the usernames ``NotableWalrus'' and ``PsyduckIsUbers'', which are not always the same model and do not appear in results, but may be present in replays or videos featured in our release materials.}
\label{tab:model_usernames}
\end{table}

\begin{figure}[h!]
    \centering
    \includegraphics[width=.9\linewidth]{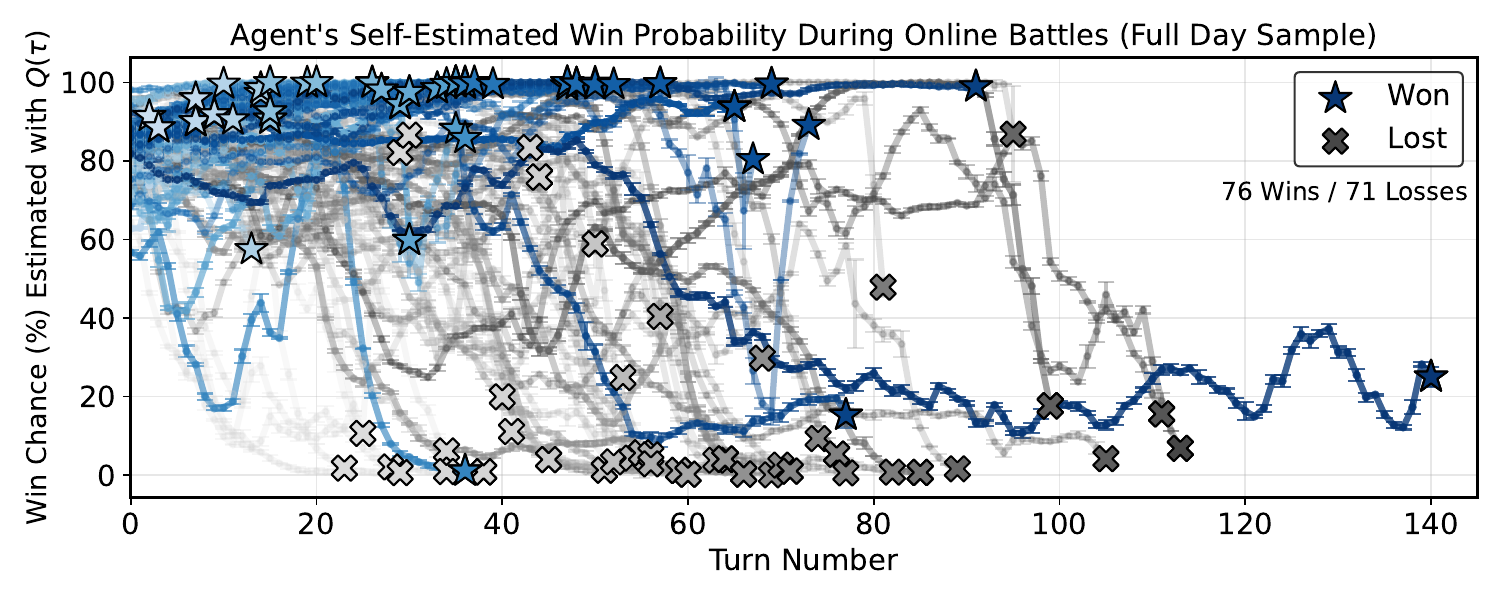}
    \caption{\textbf{$Q$-functions as a win estimate.} We track critic value predictions (for $\gamma = .999$) during battles across a $24$-hour period of the Large-RL model's gameplay on the PS ladder. If we simplify by ignoring the reward function's small shaping terms and the discount factor, we can plot these values as a more interpretable estimate of win probability. We mark these value series by their true outcome. Small error bars denote two standard deviations over the ensemble of $4$ critics.}
    \label{fig:win_prediction}
\end{figure}

We believe that the ability to generate human-like gameplay at fast inference speeds with arbitrarily prompted teams can be a fun and useful practice tool for human players. However, the models can be frustrating to play against because their reward function encourages delaying losses, and they do not forfeit. Figure \ref{fig:win_prediction} uses the Large RL model to show that Q-value predictions are calibrated enough to identify lost positions and implement auto-forfeits.

\begin{table}[]
\begin{tabular}{llcccc}
\hline
Model                              & Username                               & Gen1 OU  & Gen2 OU & Gen3 OU & Gen4 OU \\ \hline
\multicolumn{1}{l|}{PokeEnv Heuristic} & \multicolumn{1}{l|}{WinningIsOptional} & 16 - 59  & N/A     & 16 - 54 & 21 - 36 \\
\multicolumn{1}{l|}{Small IL}          & \multicolumn{1}{l|}{SmallSparks}       & 54 - 66  & 20 - 63 & 25 - 50 & 41 - 59 \\
\multicolumn{1}{l|}{Large RL}          & \multicolumn{1}{l|}{Montezuma2600}     & 72 - 60  & 49 - 56 & 68 - 67 & 32 - 42 \\
\multicolumn{1}{l|}{SynRL-V0}          & \multicolumn{1}{l|}{Metamon1}          & 57 - 42  & 42 - 35 & 61 - 52 & 49 - 51 \\
\multicolumn{1}{l|}{SynRL-V1}          & \multicolumn{1}{l|}{TheDeadlyTriad}    & 107 - 64 & 76 - 52 & 82 - 74 & 83 - 61 \\
\multicolumn{1}{l|}{SynRL-V1+SP}       & \multicolumn{1}{l|}{ABitterLesson}     & 65 - 38  & 51 - 30 & 80 - 71 & 64 - 56 \\
\multicolumn{1}{l|}{SynRL-V1++}        & \multicolumn{1}{l|}{QPrime}            & 72 - 52  & 37 - 24 & 48 - 32 & 68 - 53 \\
\multicolumn{1}{l|}{SynRL-V2}          & \multicolumn{1}{l|}{MetamonII}         & 148 - 95 & 75 - 40 & 76 - 53   & 68 - 48
\end{tabular}
\caption{\textbf{PS Usernames and Win - Loss Records.}}
\label{tbl:raw_win_loss}
\end{table}

%%%%%%%%%%%%%%%%%%%%%%%%%%%%%%%%%%%%%%%%%%%%%%%%%%%%%%%%%%%%%%%%
% AUTHOR: If your paper has no supplementary materials, you may 
%         comment out the line below, which creates the title for
%         the supplementary materials.
%%%%%%%%%%%%%%%%%%%%%%%%%%%%%%%%%%%%%%%%%%%%%%%%%%%%%%%%%%%%%%%%
%\beginSupplementaryMaterials

\end{document}